\documentclass{article} 
\usepackage{ICLR/iclr2021_conference, times}

\usepackage[utf8]{inputenc} 
\usepackage[T1]{fontenc}    
\usepackage[draft]{hyperref}       
\usepackage{url}            
\usepackage{booktabs}       
\usepackage{amsfonts}       
\usepackage{nicefrac}       
\usepackage{microtype}      
\usepackage{amsmath}
\usepackage{amsthm}
\usepackage{subcaption}
\usepackage{flushend}
\usepackage{times}
\usepackage{graphicx}
\usepackage{wrapfig,lipsum}
\usepackage{enumitem}
\usepackage{amssymb}
\usepackage{wrapfig}
\setlist{nolistsep,leftmargin=*}
\usepackage{xcolor}
\usepackage{xspace}
\usepackage{verbatim}
\usepackage{physics} 
\usepackage[compact]{titlesec}
\titlespacing{\section}{0pt}{*0}{*0}
\titlespacing{\subsection}{0pt}{*0}{*0}
\titlespacing{\subsubsection}{0pt}{*0}{*0}
\usepackage{thmtools,thm-restate}
\usepackage{adjustbox}
\DeclareMathOperator*{\argmin}{arg\,min}

\newcommand{\varun}[1]{\textcolor{blue}{Varun: #1}}

\newcommand{\cz}[1]{\textcolor{black}{#1}}

\newcommand*{\ie}{{\em i.e.,}\@\xspace}

\usepackage{color,comment,rotating,url,xspace} 
\usepackage{tikz}
\usetikzlibrary{arrows,shadows,shapes,backgrounds,decorations,snakes,fit}

\usepackage{authblk}
\usepackage{algorithmic}

\title{Causally Constrained Data Synthesis for Private Data Release}

\author[1, 2]{\small Varun Chandrasekaran}
\author[2]{\small Darren Edge}
\author[1]{\small Somesh Jha}
\author[2]{\small Amit Sharma}
\author[2]{\small Cheng Zhang}
\author[2]{\small Shruti Tople}

\affil[1]{University of Wisconsin-Madison}
\affil[2]{Microsoft Research}

\usepackage{tcolorbox}
\usepackage{multirow}
\usepackage{multicol}
\usepackage{algorithm}
\usepackage{thmtools}

\iclrfinalcopy

\begin{document}

\maketitle

\begin{abstract}

Making evidence based decisions requires data. However for real-world applications, the privacy of data is critical. Using synthetic data which reflects certain statistical properties of the original data preserves the privacy of the original data. To this end, prior works utilize differentially private data release mechanisms to provide formal privacy guarantees. However, such mechanisms have unacceptable privacy vs. utility trade-offs. We propose incorporating {\em causal information} into the training process to favorably modify the aforementioned trade-off. We theoretically prove that generative models trained with additional causal knowledge provide stronger differential privacy guarantees. Empirically, we evaluate our solution comparing different models based on variational auto-encoders (VAEs), and show that causal information improves resilience to membership inference, with improvements in downstream utility.

\end{abstract}

\section{Introduction}

\cz{Automating AI-based solutions and making evidence-based decisions both require data analyses. However, in many situations, such as healthcare~\citep{jordon2020hide} or social justice~\citep{edge2020design}, the data is very sensitive and cannot be published directly to protect the privacy of participants involved.} Synthetic data generation, which captures certain statistical properties of the original data, is useful \cz{to resolve these issues.} However, \cz{naive data creation may not work;} when improperly constructed, the synthetic data can leak information about its sensitive counterpart (from which it was constructed). Several {\em membership} and {\em attribute inference} attacks have been shown to impact specific subgroups~\citep{mukherjee2019privgan,stadler2020synthetic,zhang2020privsyn}, and eliminate any privacy advantage provided by releasing synthetic data.
\cz{Therefore, efficient privacy-preserving synthetic data generation methods are needed. }

The de-facto mechanism used for providing privacy in synthetic data release is that of differential privacy~\citep{dwork2006calibrating}. 
The privacy vs. utility trade-off inherent with these solutions is further exacerbated in tabular data because of the correlations between different records, and among different attributes within a record as well. In such settings, the amount of noise required to provide meaningful privacy guarantees often destroys any discernible signal. Apart from assumptions made on the independence of records and attributes, prior works make numerous assumptions about the nature of usage of the synthetic dataset~\citep{xiao2010differentially,hardt2010simple,cormode2019answering,dwork2009complexity}. This results in heuristic and one-off solutions that do not easily port to different settings.

To this end, we propose a mechanism to create synthetic data that is agnostic to the downstream task. Similar to prior work~\citep{jordon2018pate}, our solution involves training a generative model to provide formal differential privacy guarantees. Unlike prior solutions, however, we encode domain knowledge into the generation process to enable better utility. In particular, to induce favorable privacy vs. utility trade-offs, our main contribution involves incorporating  {\em causal information} while training synthetic data generative models. We formally prove that generative models trained with causal knowledge of the specific dataset are more private than their non-causal counterparts. Prior work has shown the privacy benefits of causal learning in only discriminative setting~\citep{tople2020alleviating}. In comparison, our result is more generic, and applies across the spectrum (for both discriminative and generative models).

\cz{Based on our theoretical results, we present a novel practical solution, utilizing } advances in \cz{deep generative models, in particular, variations autoencoder (VAE) based models~\citep{kingma2013auto,rezende2014stochastic}}. These models combine the advantage of both deep learning and probabilistic modeling, making them scale to large datasets, flexible to fit complex data in a probabilistic manner, and  can be used for data generation~\citep{ma2019eddi, ma2020vaem}. Thus, in designing our solution, we assume structured causal graphs (SCGs) as prior knowledge and train {\em causally consistent} VAEs. Additionally, we train causally consistent differentially-private VAEs to understand the benefit of causality in conjunction to that of differential privacy. 

We experimentally evaluate our solution for utility and privacy trade-off on two real world applications: (a) a medical dataset regarding neuropathetic pain~\citep{tu2019neuropathic}, and (b) a student response dataset from a real-world online education platform~\citep{wang2020diagnostic}. Based on causal knowledge provided by a domain expert, we show that models that are causally consistent are more stable by design. In the absence of differentially-private noise, causal models elevate the baseline utility by $2.42$ percentage on average while non-causal models degrade it by $3.49$ percentage. Consequently, when provided the same privacy budget, causally consistent models are more utilitarian in downstream tasks than their non-causal counterparts. This demonstrates that causal models allow us to extract more of the utility from the synthetic data while maintaining theoretical privacy guarantees.

With respect to privacy evaluation, prior works solely rely on the value of the privacy budget $\varepsilon$. We take this one step further and empirically evaluate attacks. Our experimental results demonstrate the positive impact of causality in inhibiting the membership inference adversary's advantage. Better still, we demonstrate that differentially-private models that incorporate both complete or even partial (or incomplete) causal information are more resilient to membership inference adversaries than those with purely differential privacy with the exact same $\varepsilon$-DP guarantees. These results strengthen the need for using empirical attacks as an additional measure to evaluate privacy guarantees of a given private generative model. 

In summary, the contributions of our work include:

\begin{enumerate}
\item A deeper understanding of the advantages of causality through a theoretical result that highlights the privacy amplification induced by causal consistency (\S~\ref{sec:thm}).
\item An efficient and scalable mechanism for causally-consistent and differentially private data release (\S~\ref{sec:method}). 
\item Empirical results demonstrating that causally constrained (and differentially private) models are more resilient to membership inference as compared to non-causal counterparts, without significant accuracy degradation (\S~\ref{sec:eval}).
\end{enumerate}

\section{Problem Statement \& Notation}

{\bf Problem Statement:} We study the problem of {\em private data release}, where a data owner is in possession of a sensitive dataset which they wish to release. Formally, we define a dataset $D$ to be the set $\{\mathbf{x}_1, \cdots \mathbf{x}_n\}$ of $n$ records, where each record $\mathbf{x}=(x_1, \cdots, x_k)$ has $k$ attributes. We refer to $n$ as the dataset size, and $d$ as the dimensionality of the data. Each record $\mathbf{x}$ belongs to the universe of records $\mathcal{X}$. 

Our approach involves designing a procedure which takes in a {\em sensitive dataset} $D_p$ and outputs a synthetic dataset $D_s$ which has some formal privacy \cz{guarantee and maintains properties from original dataset for downstream decision-making tasks}. Formally speaking, we wish to design $f_\theta: Z \rightarrow \mathcal{X}$, where $\theta$ are the parameters of the method, and $Z$ is some underlying latent representation for inputs in $\mathcal{X}$.

\subsection{Preliminaries}

In our work, we wish for $f_\theta$ to provide the guarantee of differential privacy. Below, we define useful concepts related to the same.

{\bf Differential Privacy ~\citep{dwork2006calibrating}:}. Let $\varepsilon \in \mathbb{R}^+$ be the privacy budget, and $H$ be a randomized mechanism that takes a dataset as input. $H$ is said to provide  $\varepsilon$-differential privacy if, for all datasets $D_1$ and $D_2$ that differ on a single record, and all subsets $S$ of the outcomes of running $H$:  $\mathbb{P}[H(D_{1})\in S]\leq e^{\varepsilon}\cdot \mathbb{P}[H(D_{2})\in S]$, where the probability is over the randomness of $H$. 

\noindent{\bf Sensitivity:} Let $d \in \mathbb{Z}^+$, ${\mathcal {D}}$ be a collection of datasets, and define $H : \mathcal{D} \rightarrow \mathbb{R}^d$. The $\ell_1$ sensitivity of $H$, denoted  $\Delta H$, is defined by $\Delta H=\max \lVert H(D_{1})-H(D_{2})\rVert _{1}$, where the maximum is over all pairs of datasets $D_{1}$ and $D_{2}$ in ${\mathcal {D}}$ differing in at most one record.

We rely on generative models to enable private data release. Formally, if the generative models are trained to provide differential privacy, then any further post-processing (such as using the generative model to obtain a {\em synthetic dataset}) is also differentially private by the post-processing property of differential privacy~\citep{dwork2014algorithmic}.

\noindent {\bf \cz{Causally Consistent Models}:} Formally, the \cz{underlying }data generating process (DGP) is characterized by a causal graph that describes the conditional independence relationships between different variables. \cz{In this work, we use the term causally consistent models to refer to those models that factorize in the causal direction. }
For example, the graph $X_1 \rightarrow X_2$ implies that the 
factorization \cz{following the causal direction is $\mathbb{P}(X_1, X_2)= \mathbb{P}(X_1) \cdot \mathbb{P}(X_2|X_1)$}.
\cz{Due to the {\em modularity property}~\citep{woodward2005making}, the mechanism to generate $x_2$ from $x_1$ is independent from the marginal distribution of $x_1$. This only holds in causal factorization but not in anti-causal factorization.}

\section{Privacy Amplification Through Causality}
\label{sec:thm}

In this section, we present our main result: {\em causally consistent (or causal) models are more private than their non-causal/associational counterparts.} Our focus is on generative models, but our analysis is applicable to both generative and discriminative contexts. 

\subsection{Main Theorem}
\label{sec:theorem}

 We begin by introducing notations and definitions needed for our proof. A mechanism $M$ takes in as input a dataset $D$ and outputs a parameterized model $f_\theta$. The model output by the mechanism $M$ belongs to a hypothesis space $\mathcal{H}$. The dataset $D$ comprises of samples, where each sample $\mathbf{x}=(x_1, \cdots, x_k)$ comprises of $k$ features. To learn the model, we utilize {\it empirical risk mechanism (ERM)}, and assume a loss function $\mathcal{L}$ that
is Lipschitz continuous and strongly convex. $\mathcal{L}_D$ denotes the average loss calculated over the dataset $D$ (i.e., $\mathcal{L}_D = \frac{1}{n} \sum_{\mathbf{x} \in D} \mathcal{L}_{\mathbf{x}}$). $\mathcal{L}_\mathbf{x}$ denotes the loss being calculated over sample $\mathbf{x}$ i.e., $\mathcal{L}_\mathbf{x}(f_\theta) = \mathcal{L}_\mathbf{x}(f_\theta(\mathbf{x}))$.
For generative models, the loss should reflect how well the model fits the data while for probabilistic models, and if we use variational inference,
we can not minimize the KL divergence exactly, but one instead minimizes a function that is equal to it up to a constant, which is the evidence lower bound (ELBO).

\noindent{\bf Data Generating Process (DGP):}  DGP $<f^*, \eta>$ is obtained as follows: $f^* = \lim_{n \to \infty} \argmin \mathcal{L}_{D}(f_\theta)$. Essentially $f^\star$
can be thought of as the {\em infinite data} limit
of the ERM learner and can be viewed 
as the {\em ground truth}. In a causal setting, the DGP for a variable $X$ is defined as $f^*(X) = (f^*_1(Pa(X_1)) + \eta_i, \cdots, f^*_n(Pa(X_n))+\eta_n)$ where $\eta_i$ are mutually, independently chosen noise and $Pa(X_i)$ are the parents
of $X_i$ in the SCG.

\noindent{\bf Loss-maximizing (LM) Adversary}: Given a model $f_\theta$ and a loss function $\mathcal{L}$, an LM adversary chooses a point $\mathbf{x}'$ (to be added to the 
dataset $D$ to obtain a neighboring dataset $D'$) as $\mathbf{x'} = \arg \max_{\mathbf{x}} \mathcal{L}_\mathbf{x}(f_\theta)$. 

\noindent{\bf Empirical Risk Minimization (ERM):}  For a given hypothesis class $\mathcal{H}$, and a dataset $D$, the ERM learner uses the ERM rule to choose parameters $\theta \in \mathcal{H}$, with the lowest possible error over $D$. Formally, $f_{\theta^*} = \argmin_{\theta \in \mathcal{H}} \mathcal{L}_D(f_\theta)$ denotes the optimal model.

\begin{restatable}{theorem}{privacythm}
Given a dataset $D$ of size $n$, and a  strongly convex and Lipschitz continuous loss function $\mathcal{L}$, assume we train two models in a differentially private manner: a causal (generative) model $f_{\theta_c }$, and an associational (generative) model $f_{\theta_a}$, such that they minimize $\mathcal{L}$ on $D$. Assume that the class of hypotheses $\mathcal{H}$ is expressive enough such that the true causal function lies in $\mathcal{H}$.
\begin{enumerate}
\itemsep0em
    \item {\it Infinite sample case.} As $n \to \infty$, the privacy budget of the causal model is lower than that of the associational model \ie $\varepsilon_c \leq \varepsilon_a$.
    \item {\it Finite sample case.}  For finite $n$, assuming certain conditions on the associational models learnt, the privacy budget of the causal model is lower than that of the associational model \ie $\varepsilon_c \leq \varepsilon_a$.
\end{enumerate}
\end{restatable}

\noindent{\bf Proof Outline:} The detailed proof is found in Appendix~\ref{sec:proof}. The main steps  of our proof
are as follows:
\begin{enumerate}
\item We show that the maximum loss of a causal model is lower than or equal to maximum loss of the corresponding associational model. 
\item Using strong convexity and Lipschitz continuity of the loss function, we show how the difference in loss corresponds to the sensitivity of the learning function.
\item Finally, knowledge of sensitivity directly helps determine the privacy budget $\varepsilon$. 

\end{enumerate}

We present a key idea behind  step 1 of the proof
of our main theorem. 
Assume an LM adversary and a strongly-convex loss function $\mathcal{L}$. Given a causal model $ f_{\theta_c}$ and an associational model $f_{\theta_a}$ trained on dataset $D$ using ERM; the LM adversary selects two points: $\mathbf{x}'$ and $\mathbf{x}''$. We wish to show that the  {\em worst-case loss} obtained on the causal ERM model $\mathcal{L}_{\mathbf{x}'}(f_{\theta_c})$ is lower than the worst-case loss obtained on the associational ERM model $\mathcal{L}_{\mathbf{x}''}(f_{\theta_a})$, or
\begin{equation}
    \max_{\mathbf{x}} \mathcal{L}_{\mathbf{x}}(f_{\theta_c}) \leq \max_{\mathbf{x}} \mathcal{L}_{\mathbf{x}}( f_{\theta_a})
\end{equation}

We provide intuition for the equation given above
in the infinite data case. The case of finite data appears in Appendix~\ref{sec:proof}. The key idea is that {\it in the causal model, an LM adversary is constrained by the causal structure, but does not have these constraints in the associational setting.}

In the infinite data case,~\citet{peters2017elements} demonstrate that, for causal models 
$\lim_{n \to \infty} \mathcal{L}_D (f_{\theta_c}) = \lim_{n \to \infty} \arg \max \mathcal{L}_D (f_\theta) = f^*$.
In other words, in the causal setting an LM adversary is 
constrained by the DGP $f^\star$. However, the same is not true for associational models (i.e.,  $\lim_{n \to \infty} \mathcal{L}_D (f_{\theta_a})  \neq f^*$). In the causal world
LM adversary is more constrained than in the associational world. Thus, 
\begin{equation*}
    \max_{\mathbf{x}} \mathcal{L}_{\mathbf{x}}(f_{\theta_c}) = \eta \leq \max_{\mathbf{x}} \mathcal{L}_{\mathbf{x}}( f_{\theta_a})
\end{equation*}

Once this is established in the finite data regime as well, we will utilize specific properties of the loss function to bound the sensitivity of the learned parameters $\theta_c$ and $\theta_a$ and use this information to obtain bounds on $\varepsilon_c$ (the privacy budget of the causal model) and $\varepsilon_a$ (the privacy budget of the associational model).

\section{Causal Deep Generative Models}
\label{sec:method}

\cz{As we aim to release a privacy-preserving synthetic dataset, we require a generative model for data generation.
Based on our theoretical analysis and considering the flexibility and computational efficiency requirement in large-scale real-world applications, we use causally consistent deep generative models. In particular, our solution is based on Variational Auto-Encoders (VAEs). }

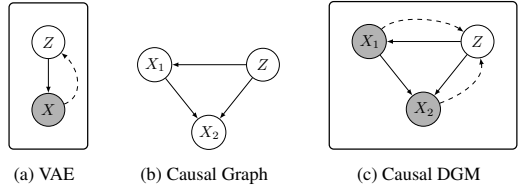
\begin{figure}[h]
\centering
\subfloat[VAE]{\scalebox{0.6}{\pgfdeclarelayer{background}
\pgfdeclarelayer{foreground}
\pgfsetlayers{background,main,foreground}

\begin{tikzpicture}

\tikzstyle{scalarnode} = [circle, draw, fill=white!11,  
    text width=1.2em, text badly centered, inner sep=2.5pt]
\tikzstyle{arrowline} = [draw,color=black, -latex]
;
\tikzstyle{dasharrowline} = [draw,dashed, color=black, -latex]
;
\tikzstyle{surround} = [thick,draw=black,rounded corners=1mm]
    \node [scalarnode, fill=black!30] at (0,0) (X)   {$X$};

    \node [scalarnode] at (0, 1.5) (Z) {$Z$};

    \path [arrowline]  (Z) to (X);


    \path [dasharrowline]  (X) to[out=20,in=-40, distance=0.5cm ] (Z);

    \node[surround, inner sep = .5cm] (f_N) [fit = (Z)(X) ] {};
\end{tikzpicture}}}
\hspace{10pt}
\subfloat[Causal Graph]{\scalebox{0.6}{\pgfdeclarelayer{background}
\pgfdeclarelayer{foreground}
\pgfsetlayers{background,main,foreground}

\begin{tikzpicture}

\tikzstyle{scalarnode} = [circle, draw, fill=white!11,  
    text width=1.2em, text badly centered, inner sep=2.5pt]
\tikzstyle{arrowline} = [draw,color=black, -latex]
;
    \node [scalarnode] at (0,0) (X)   {$X_2$};
    \node [scalarnode] at (-1.2, 1.5) (Y) {$X_1$};
    \node [scalarnode] at (1.2, 1.5) (Z) {$Z$};
    \path [arrowline]  (Y) to (X);
    \path [arrowline]  (Z) to (X);
    \path [arrowline]  (Z) to (Y);
\end{tikzpicture}}}
\hspace{10pt}
\subfloat[Causal DGM]{\scalebox{0.6}{\pgfdeclarelayer{background}
\pgfdeclarelayer{foreground}
\pgfsetlayers{background,main,foreground}

\begin{tikzpicture}

\tikzstyle{scalarnode} = [circle, draw, fill=white!11,  
    text width=1.2em, text badly centered, inner sep=2.5pt]
\tikzstyle{arrowline} = [draw,color=black, -latex]
;
\tikzstyle{dasharrowline} = [draw,dashed, color=black, -latex]
;
\tikzstyle{surround} = [thick,draw=black,rounded corners=1mm]
    \node [scalarnode, fill=black!30] at (0,0) (X)   {$X_2$};
    \node [scalarnode, fill=black!30] at (-1.2, 1.5) (Y) {$X_1$};
    \node [scalarnode] at (1.2, 1.5) (Z) {$Z$};
    \path [arrowline]  (Y) to (X);
    \path [arrowline]  (Z) to (X);
    \path [arrowline]  (Z) to (Y);


    \path [dasharrowline]  (X) to[out=20,in=-80, distance=0.5cm ] (Z);
    \path [dasharrowline]  (Y) to[out=45,in=145, distance=0.5cm ] (Z);

    \node[surround, inner sep = .5cm] (f_N) [fit = (Z)(X)(Y) ] {};
\end{tikzpicture}}}
 \caption{\cz{Exemplar case comparing our solution to VAE.} } 
\label{fig:overview}
\end{figure}

\noindent{\bf VAE Recap:} In VAE~\citep{kingma2013auto,rezende2014stochastic}, the data generation, $p_\theta (\mathbf{x}|z)$, is realized by a deep neural network (and is hence parameterized by $\theta$) known as the {\em decoder}. To approximate the posterior of the latent variable $p_\theta (z|\mathbf{x})$, VAEs use another neural network (the {\em encoder}) with $\mathbf{x}$ as input to produce an approximation of the posterior $q_\phi(z|\mathbf{x})$. VAEs are trained by maximizing an evidence lower bound (ELBO), which is equivalent to minimizing the KL divergence between $q_\phi(z|\mathbf{x} )$ and $p_\theta (z|\mathbf{x})$~\citep{jordan1999introduction,wainwright2008graphical,blei2017variational,zhang2018advances}. \cz{Naive solution involving the use of VAE for data generation would concatenate all variables as $X$, train the model, and generate data through sampling from the prior $p(Z)$.} To train the model, we wish to minimize the KL divergence between the true posterior $p(z|\mathbf{x})$ and the approximated posterior $q_\phi(z|\mathbf{x})$. This is achieved by maximizing the ELBO defined as:
$$\text{ELBO}=\mathbb{E}_{q_\phi(z|\mathbf{x}))}[\log p_\theta(\mathbf{x}|z)] - \mathbb{K}\mathbb{L}[q_\phi((z|\mathbf{x})||p(z)]$$

\noindent{\bf Causal Deep Generative Models:} 
Now that we have established the elements required to build our solution, we provide a brief overview. We wish to learn a differentially private generative model of the form $f_\theta(Z) = \cal{X}$. To this end, we wish to incorporate causal properties associated with the distribution of inputs. \cz{VAEs are probabilistic graphical models, and provide a way to design the data generation process for any joint probability distribution. Here, we propose to make the model generative process consistent with the given causal relationship.} Figure~\ref{fig:overview} \cz{provides an example of how this is achieved. As an example, the original dataset contains variable $X_1$ and $X_2$ and the causal relationship follows Figure~\ref{fig:overview} (b). In this case, $X_1$ can be a medical treatment and $X_2$ can be a medical test, and $Z$ is the patient health status which is not observed. Instead of using VAE to generate the data, we design a generative model as shown in Figure~\ref{fig:overview} (c), where the solid line shows the model $p(x_1,x_2,z)=p(z)p_{\theta_1}(x_1|z)p_{\theta_2}(x_2|x_1,z)$, and the dashed line shows the inference network $q_\phi(z|x_1,x_2)$. In this way, the model is consistent with the underlying causal graph. The modeling principle is similar to that of CAMA~\citep{zhang2020causal}. However, CAMA only focuses on prediction task and ignores all variables out of the Markov Blanket of the target. In our application, we aim for data generation and need to consider {\em the full causal graph}. In this sense, our proposed solution examples generalizes CAMA. }

\noindent{\bf Why does causality provide any benefit?} Any dataset can be thought of as data collected from multiple distributions. In the case of causal models, we know the proper factorization of the joint probability distribution, and any model we learn to fit these factorized distributions will be more {\em stable}. Additionally, the model is well specified based on these factorizations. In the associational case, the model is capable of learning any kind of relationship between different attributes, even those that may not be stable. Due to the higher stability in the causal model, its (worst-case) loss on unseen points (or points generated by the true DGP) is lower.
As we show in our theorem in \S~\ref{sec:theorem}, this translates to lower sensitivity which results to better privacy. 

Put another way, associational models may observe certain correlations (useful or spurious) in certain points in the dataset and learn to fit to them. These models may consequently not learn from other points in the dataset. However, this phenomenon will not happen in causal models.

\noindent{\bf Remark:} \cz{In this work, we assume that the causal relationship is given. In practice, this can be obtained from domain expert or using a careful chosen causal discovery algorithm \cite{glymour2019review,tu2019causal,spirtes2000causation}.  }

\section{Implementation}

We describe important features related to our implementation. Refer Appendix~\ref{app:training} for more details.

\subsection{Code \& Datasets}
\label{sec:code}

As part of our experiments, we utilize datasets \cz{from two real-world applications. The first one is }  the \texttt{EEDI} dataset~\citep{wang2020diagnostic} \cz{which is one of the largest real-world education data collected from an online education platform. It} contains answers by students (of various educational backgrounds) for certain diagnostic questions. \cz{The second one is} the neuropathic pain (\texttt{Pain}) diagnosis dataset obtained from a causally grounded simulator~\citep{tu2019neuropathic}. For this dataset, we consider two variants: one with 1000 data records (or \texttt{Pain1000}), and another with 5000 data records (or \texttt{Pain5000}). More salient features of each dataset is presented in Table~\ref{tab:salient}. We choose these datasets as they encompass diversity in their size, dimensionality $k$, and have some prior information on causal structures (refer Appendix~\ref{sec:graphs} for more details). We utilize this causal information in building a causally consistent generative model. \cz{The (partial) causal graph is given utilizing domain knowledge in these two contexts.} 

\begin{table}[t]
\centering
\small
\begin{tabular}{@{}cccccc@{}}
\toprule
 \bf Model &    \bf Dataset               &    \bf \# Records             & $k$ & $C$ &  $\varepsilon$ \\ \midrule \midrule
 Causal & \multirow{2}{*}{\texttt{EEDI}} & \multirow{2}{*}{2950} & \multirow{2}{*}{948} & 1.35 & \multirow{2}{*}{12.42} \\
 Non Causal &                   &                   &  & 1.07 \\ \hline
 Causal & \multirow{2}{*}{\texttt{Pain5000}} & \multirow{2}{*}{5000} & \multirow{2}{*}{222} & 0.17 & \multirow{2}{*}{5.62} \\
 Non Causal &                   &                   &  & 0.33 \\ \midrule
 Causal & \multirow{2}{*}{\texttt{Pain1000}} & \multirow{2}{*}{1000} & \multirow{2}{*}{222} & 0.25  & \multirow{2}{*}{2.36} \\
 Non Causal &                   &                   &  &  0.41\\ \midrule
 Causal & \multirow{2}{*}{\texttt{Synthetic}} & \multirow{2}{*}{1000} & \multirow{2}{*}{22} &  0.55 & \multirow{2}{*}{3.9} \\
 Non Causal &                   &                   &  & 0.65 \\ \bottomrule
\end{tabular}
\caption{Salient features of our experimental setup. Parameters required for DP training are located in Appendix~\ref{app:training}.}
\label{tab:salient}
\end{table}

We performed all our experiments on a server with 8 NVidia GeForce RTX GPUs, 48 vCPU cores, and 252 GB of memory. All our code was implemented in \texttt{python}. \cz{As the \texttt{EEDI} dataset has missing data-entries, we} utilize the \cz{partial VAE techniques specified in the work of~\citet{ma2019eddi} in our causal consistent model to handle such heterogeneous observations.} Our code is released as part of the supplementary material.

\subsection{Privacy Budget ($\varepsilon$)} 
For training our differentially private models, we utilize \texttt{opacus v0.10.0} library\footnote{\url{https://github.com/pytorch/opacus}} that supports the DP-SGD training approach proposed by~\citet{abadi2016deep} of clipping the gradients and adding noise while training.
We ensure that the training parameters for training both causal and non-causal models are fixed. These are described in Appendix~\ref{app:training}. 
 For all our experiments, we perform a grid search over the space of {\em clipping norm $C$} and {\em noise multiplier $\sigma$}.
Once training is done, we calculate the privacy budget after training using the Renyi differential privacy accountant provided as part of the \texttt{opacus} package. The reported value of $\varepsilon$ for both causal and non-causal models is the same (and can be found in Table~\ref{tab:salient}). 

\subsection{Membership Inference Attack} 
\label{sec:mi}

Prior solutions for private generative models often use the value of $\varepsilon$ as the sole measure for privacy~\citep{zhang2017privbayes,jordon2018pate,zhang2020privsyn}. In addition to computing $\varepsilon$, we go a step further and use a membership inference (MI) attack specific to generative models to empirically evaluate if the models we train leak information~\citep{stadler2020synthetic}. In this attack, the adversary has access to (a) synthetic data sampled from a generative model trained {\em with a particular record} in the training data (line 11), and (b) synthetic data sampled from a generative model trained {\em without the same record} in the training data (line 5). The objective of the adversary (refer Algorithm~\ref{alg}) is to use this synthetic data (from both cases, as in lines 7, 13) and learn a classifier to determine if a particular record was used during training or not. Note that the notation $S \sim f^s$ implies that a dataset $S$ of size $s$ is sampled from the generative model. The various parameters associated with this approach include (a) $n_T$: the number of candidate targets considered, (b) $n$: the number of times a particular generative model is trained, (c) $t$: the size of the training data, (d) $n_s$: the number of samples obtained from the trained generative model, and (e) $s$: the size of the sample obtained from the trained generative model. Observe that the final data obtained for the attack is of the order $n_T \times n \times s$, and the number of unique generative models trained for this approach is equal to $n \times n_T \times 2$. 

Once the adversary's training data $D_{test}$ is obtained, the adversary can use several feature extractors to extract more information from the entries in $D_{test}$. As in the original work, we utilize a \texttt{Naive} feature extractor which extracts summary statistics, a \texttt{Histogram} feature extractor that contains marginal frequency counts of each attribute, a \texttt{Correlations} feature extractor that encodes pairwise attribute correlations, and an \texttt{Ensemble} feature extractor that encompasses the aforementioned extractors collectively. As part of our evaluation, we set $n_T=5$, $n=5$, $t=|D|$ for each dataset $D$, $n_s=100$, and $s=100$.

\begin{algorithm}[tb]
\footnotesize
\caption{ MI Attack~\citep{stadler2020synthetic}}
\label{alg}
\begin{algorithmic}[1]
   \STATE {\bfseries Input:} Training dataset $D$, generative model training mechanism $GM(.)$
\FOR{$i=1 \cdots n_T$}
\STATE Choose random target $\mathbf{t}$ from $D$
\FOR{$j=1 \cdots n$}
\STATE Train model $f_{out} = GM(D-\{\mathbf{t}\})$
\FOR{$k=1 \cdots n_s$}
\STATE $S_{out} \sim f_{out}^s $
\STATE $D_{test} = D_{test} \cup S_{out}$
\STATE $l_{test} = l_{test} \cup \{0\}$
\ENDFOR

\STATE Train model $f_{in} = GM(D)$
\FOR{$k=1 \cdots n_s$}
\STATE $S_{in} \sim f_{in}^s $
\STATE $D_{test} = D_{test} \cup S_{in}$
\STATE $l_{test} = l_{test} \cup \{1\}$
\ENDFOR

\ENDFOR
\ENDFOR
\STATE Train discriminative classifier using $D_{test}, l_{test}$
\end{algorithmic}
\end{algorithm}

\subsection{Utility Metrics}
Utility is preserved if the synthetic data performs equally well as the original sensitive data for any given task using any classifier. To measure the utility change with our proposed approach, we perform the following experiment: if a dataset has $k$ attributes, we utilize $k-1$ attributes to predict the $k^{th}$ attribute. We randomly choose 20 different attributes to be predicted. Furthermore, we train 5 different classifiers for this task, and compare the predictive capabilities of these classifiers when trained on (a) the original sensitive dataset, and (b) the synthetically generated private dataset. The 5 classifiers are: (a) linear SVC (or \texttt{kernel}), (b) \texttt{svc}, (c) \texttt{logistic} regression, (d) \texttt{rf} (or random forest), (e) \texttt{knn}.

Additionally, we draw pairplots using the features from the original and the synthetic dataset to compare their similarity visually. These pairplots are obtained by choosing 10 random attributes (out of the $k$ available attributes).

\section{Evaluation}
\label{sec:eval}

\begin{table*}
\tiny
\centering
\begin{tabular}{@{}cc|ccccc|ccccc@{}}
\toprule
                 \bf Dataset (Utility Range) & \bf DP & \multicolumn{5}{c|}{\bf Non Causal} & \multicolumn{5}{c}{\bf Causal} \\ \midrule \midrule
\multicolumn{2}{l|}{} &  \texttt{kernel}  & \texttt{svc}   & \texttt{logistic}    & \texttt{rf}  & \texttt{knn} & \texttt{kernel}   & \texttt{svc}    & \texttt{logistic}    & \texttt{rf}  & \texttt{knn} \\ \cmidrule{3-12}
\multirow{2}{*}{\texttt{EEDI} (86-92\%)} & $\checkmark$ &  6.83 &  7.11 & 6.82 & 6.54 & 5.36  &  -4.04  & -0.22   & -3.49   & 6.09  & 5.87  \\
                  &  $\times$ & 1.14  & 3.86  &  2.6 & 5.27 & 2.41  &  -6.83  & -2.74   & -6.13   & -0.03  & -1.96  \\ \midrule
\multirow{2}{*}{\texttt{Pain1000} (88-95\%)} & $\checkmark$ & 9.56   & 6.34    & 2.48   & 5.86  & 2.03  &  -1.84 &  4.03  &  -1.27  & 2.22  & -4.64  \\ 

                  & $\times$ &  5.42  & 5.5   & 1.31   & 6.22  & -0.924  &  -1.73 &  1.56  &  -3.16  & 2.11  & -6.92   \\ \midrule 
\multirow{2}{*}{\texttt{Pain5000} (92-98\%)} & $\checkmark$ &  4.09  &  6.89  & 6.8   & 2.27  & 5.22  & 1.7   & 2.11   &  4.58  & 0.62  & -0.82  \\
                  & $\times$ &  3.53  & 5.93   &  5.68  &  0.48 &  4.07 & -0.64   &  0.07  &  -0.24  & -4.62  & -5.13 \\  \bottomrule
\end{tabular}
\caption{{\bf Downstream Utility Change:} We report the utility change induced by synthetic data on downstream classification tasks in comparison to the original data i.e., (original data utility - synthetic data utility). Negative values indicate the percentage point improvement, while positive values indicate degradation. The performance range of the classifiers we consider is reported in parentheses next to each dataset. Observe that (a) differentially private training induces performance degradation in both causal and non-causal settings, and (b) performance degradation in the causal setting is {\em lower} than that of the non-causal setting.}
\label{tab:utility}
\end{table*}

Thus far, our discussion has focused om how (perfect) causal information can be used to theoretically minimize the privacy budget. We design our evaluation with the goal to answer the following questions:

\begin{enumerate}
\item Does the synthetic data degrade downstream utility substantially? Is this particularly true in the causal case where the privacy is amplified?
\item What is the effect of causality on MI attack accuracy when both complete and only partial causal information is available?
\item What is the effect on MI attack accuracy when a causally consitent VAE is trained both with and without differential privacy (DP)?
\end{enumerate}

We summarize our key results below:
\begin{enumerate}
\item The downstream utility is minimally affected by incorporating causality into the training procedure. In some cases, causal information increases the utility (\S~\ref{sec:utility-eval}).
\item Knowledge of a complete causal graph with differential privacy consistently reduces the adversary's advantage across different feature extractors and classifiers i.e., provides better privacy guarantees (\S~\ref{sec:privacy-eval-true}). 
\item Even partial causal information reduces the advantage of the adversary when the model is trained without DP and in majority cases with DP as well.
As a surprising result, we observe the advantage to slightly increase for specific feature extractors-dataset combination for a model trained with causality and DP (\S~\ref{sec:privacy-eval-partial}).
\end{enumerate}

\subsection{Utility Evaluation}
\label{sec:utility-eval}


\noindent{\bf Downstream Classification:} 
Table~\ref{tab:utility} shows the change in utility on different downstream tasks using the generated synthetic data when trained with and without causality as well as differential privacy. The negative values in the table indicate an improvement in utility. We only present the range of absolute utility values when trained using the original data in the table and provide individual utility for each classifier in Appendix~\ref{app:utility}.

We observe an average performance degradation of 3.49 percentage points across all non-causal models trained without differential privacy, and an average {\bf increase of 2.42} percentage point in their causal counterparts. However, it is well understood that differentially private training induces a privacy vs. utility trade-off, and consequently the utility suffers (compare rows with DP and without DP). However, when causal information is incorporated into the generative model, we observe that the utility degradation is less severe (compare pairs of cells with and without causal information). These results suggest that causal information encoded into the generative process improves the privacy vs. utility trade-off i.e., for the same $\varepsilon$-DP guarantees, the utility for causal models is better than their non-causal counterparts. 

These results are further supported by the pairplots presented in Appendix~\ref{app:utility}. They suggest that the synthetic data generated resembles the original private data.

\subsection{MI Attack Evaluation}

We train 2 generative models: one that encodes information from a SCG and one that does not. For each of these models, we train them with and without differential privacy and thus have 4 models in total for each dataset. For all our datasets, we evaluate these models against the MI adversary (\S~\ref{sec:mi}), and plot the {\em change in the adversary's advantage} which is the difference in the attack accuracy when a method (differential privacy/causal consistency) is \underline{not used} in comparison to when it is used. A positive value means that corresponding method (i.e., causality or differential privacy or both together) is effective in defending against the MI attack in that particular type of model; this consequently reduces the attacker's advantage. Note that as part of the MI attack, we are unable to utilize the \texttt{Correlation} and \texttt{Ensemble} feature extractors for the \texttt{EEDI} dataset due to computational constraints in our server.

\subsubsection{With Complete Causal Graph}
\label{sec:privacy-eval-true}

\begin{figure}
\centering
\subfloat{\includegraphics[width=\linewidth]{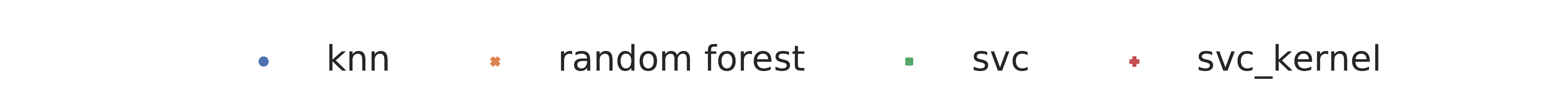}}
\vspace{-2mm}
\centering
\subfloat[{{\small Causal}}]{\label{fig:custom-full-causal}\includegraphics[width=0.5\linewidth]{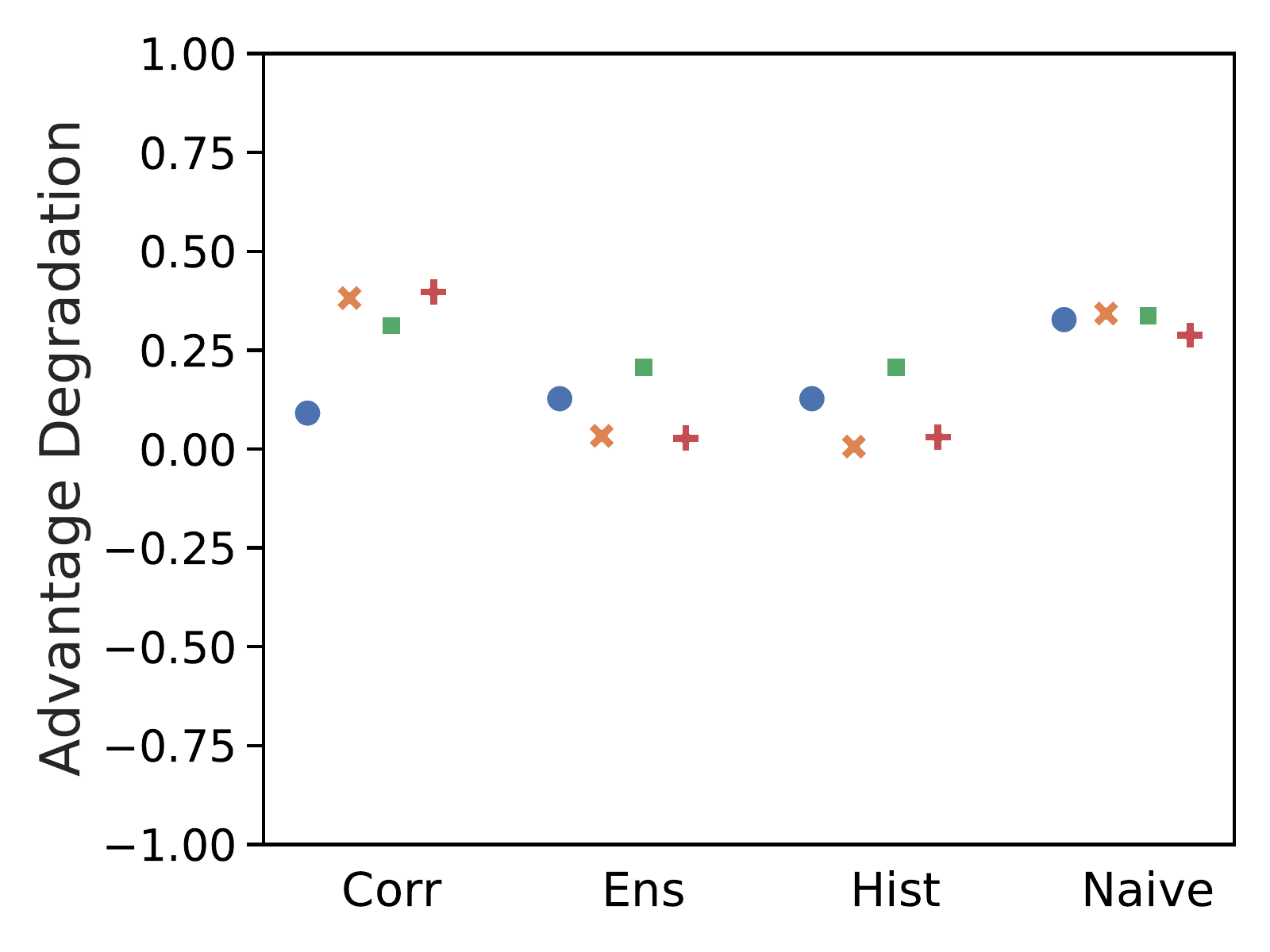}}   
\subfloat[{{\small Non Causal}}]{\label{fig:custom-full-non}\includegraphics[width=0.5\linewidth]{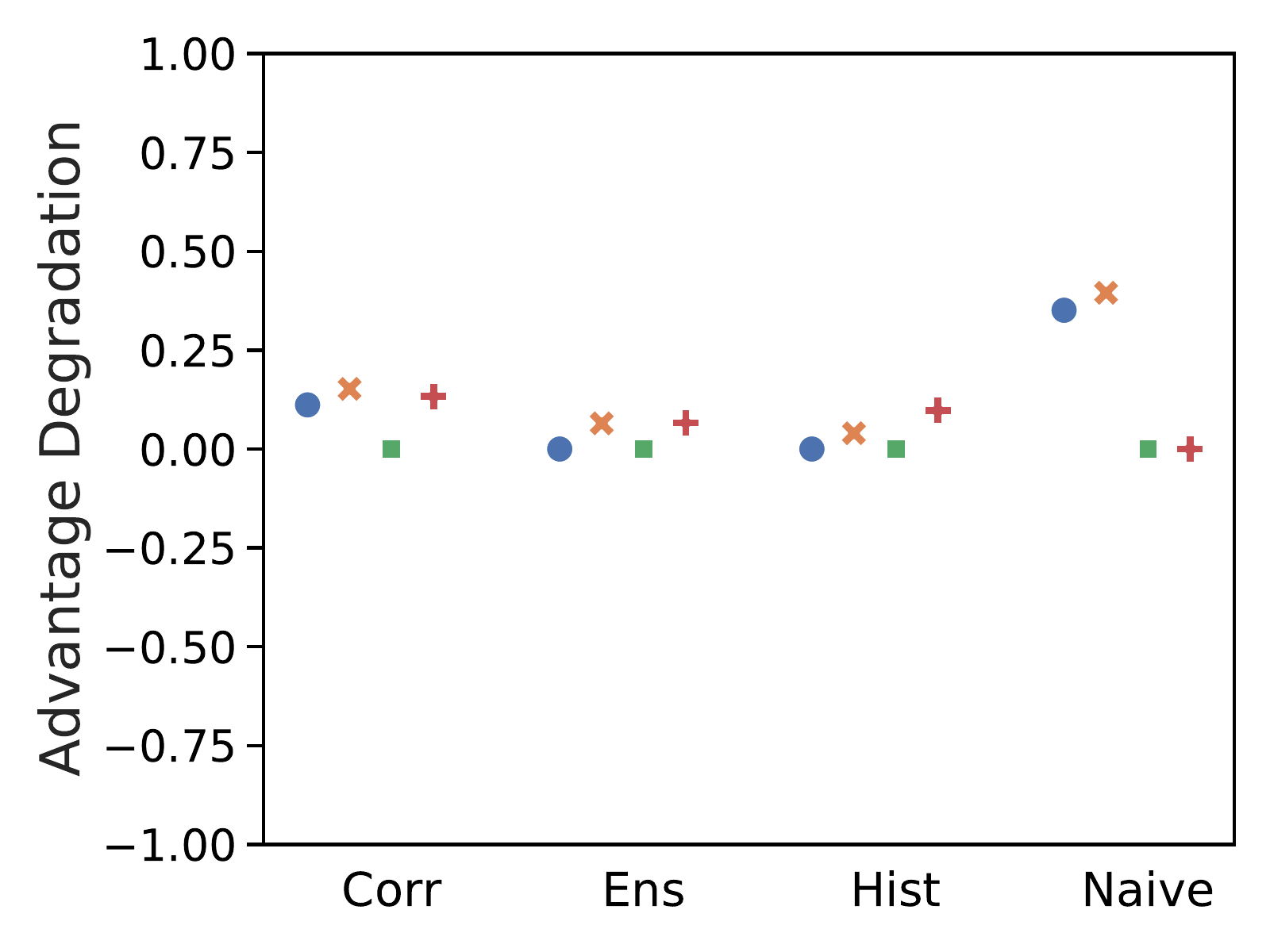}}   
\caption{{\bf True SCG:} Observe that both causal and non-causal models trained with DP reduce the adversary's advantage. Causal models provide more advantage degradation on average.} 
\label{fig:custom-full}
\end{figure}

We conduct a {\em toy experiment} with synthetic data where the complete (true) causal graph is known apriori. The data is generated based on the SCG defined in Appendix~\ref{sec:graphs}. The results are presented in Figure~\ref{fig:custom-full}. Observe that in both the causal and non-causal model, training with DP provides an advantage against the MI adversary, though to varying degrees. We also observe that causal models provide {\em greater} resilience on average in comparison to the non-causal model. It is also important to note that the reported value of $\varepsilon$ for the causal model is the same as that of the associational model. Yet, it provides better privacy guarantees. These results confirm our theoretical claim from \S~\ref{sec:theorem}: {\em perfect} causal information incorporated with DP provides boosted privacy, even against against real world adversaries. Our results also demonstrate that $\varepsilon$ cannot be used as a sole measure to gauge the privacy of a generative model. The theoretical guarantees of $\varepsilon$-DP do not always translate to resilience against membership attacks in practice. 

\subsubsection{With Partial Causal Graph}
\label{sec:privacy-eval-partial}

Many real-world datasets do not come with their associated SCGs. Learning these graphs is also a computationally expensive process. To this end, we utilize information from domain experts to partially construct a causal graph. Recall that our theoretical insight implicitly assumes that the information contained by the causal graph is holisitc and accurate. Our evaluation in this subsection verifies if the theory holds when the assumptions are violated. By partial, we mean that nodes for several variables are clubbed together to reduce the overall size of the SCG. Increasing the size of the SCG increases complexity associated with training models; resolving these issues is orthogonal to our work. 
Note that due to space constraints, we only report results for 2 out of the 3 datasets we evaluate. The trends from \texttt{Pain1000} are similar to that of \texttt{Pain5000} and are omitted for brevity. These results can be found in Appendix~\ref{app:pain}.

\begin{figure}
\centering
\subfloat{\includegraphics[width=\linewidth]{Figures/new/legend.pdf}}
\vspace{-2mm}
\centering
\subfloat[{{\small \texttt{EEDI}}}]{\label{}\includegraphics[width=0.49\linewidth]{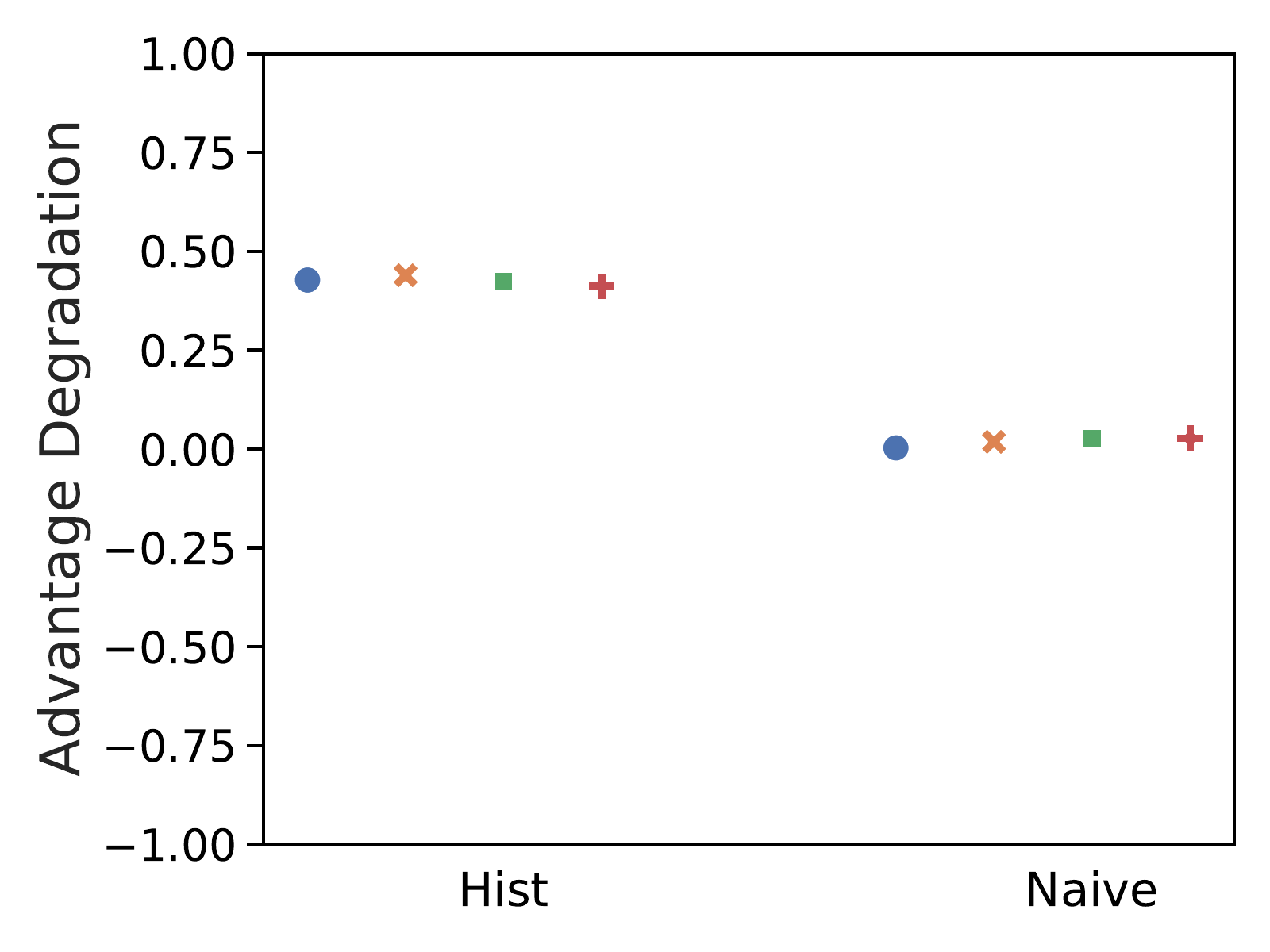}}   
\subfloat[{{\small \texttt{Pain5000}}}]{\label{}\includegraphics[width=0.49\linewidth]{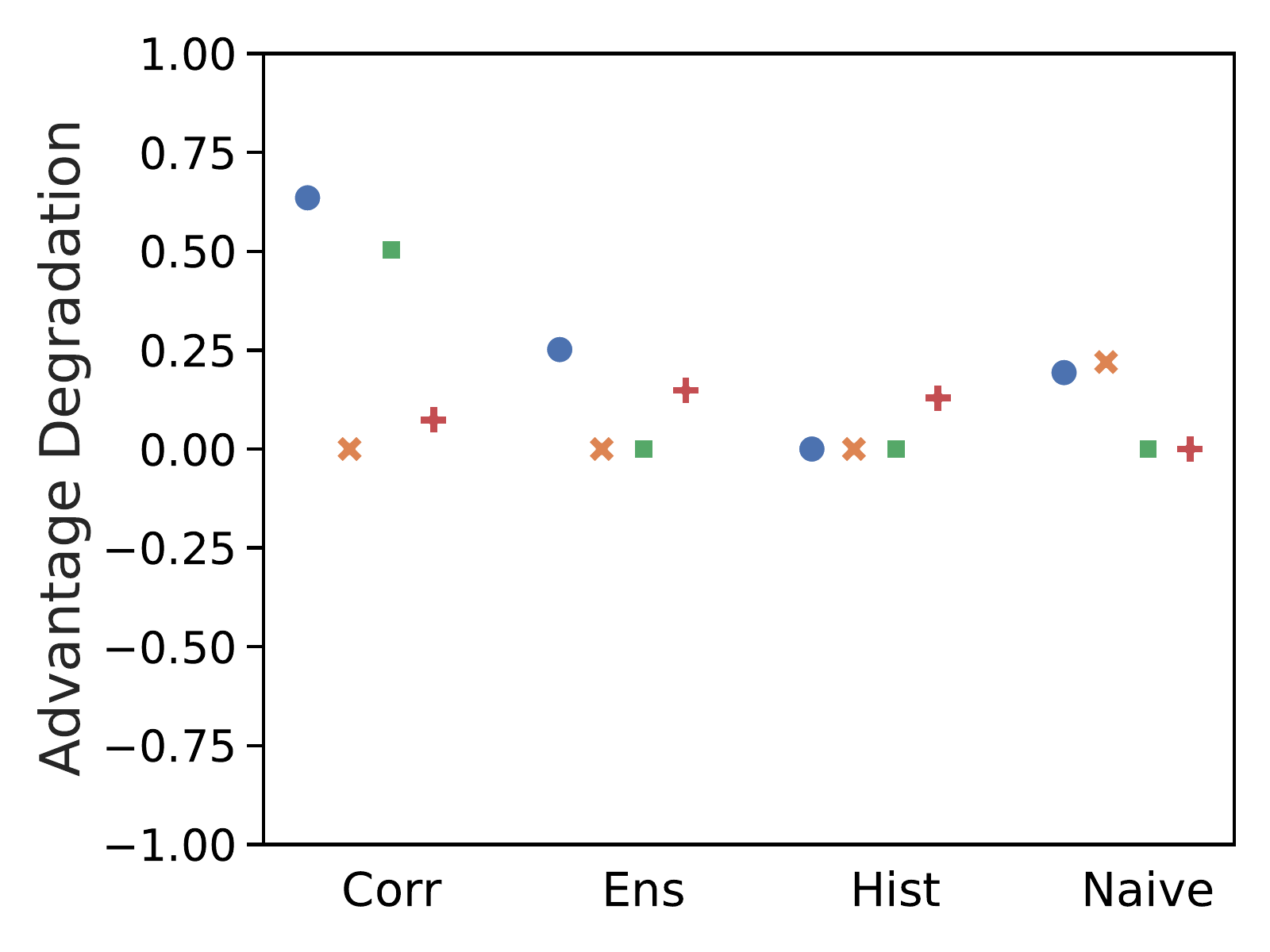}}   
\caption{{\bf Effect of DP, No Causality}: We plot the advantage degradation when the adversary uses a non-DP trained model and switches to its DP counterpart. Observe that DP enhances the defender's resilience.  } 
\label{fig:dp-effect}
\end{figure}

\paragraph{1. Effect of DP.} Figure~\ref{fig:dp-effect} plots the influence of differential private training on the MI adversary. In particular, we plot how the adversary's advantage changes when the defender switches from a model trained without DP to one that is trained with DP. A positive value indicates that DP model is more prohibitive to the adversary. We observe that the adversary's advantage is reduced to varying degrees across both datasets. This suggests that DP training is a reliable defense against adversary's of this nature~\citep{mukherjee2019privgan,bhowmick2018protection}, despite having a higher range  of $\varepsilon$ values (as shown in Table~\ref{tab:salient}). The detailed MI accuracy numbers with precision and recall are in Appendix~\ref{app:mi}.

\begin{figure}[ht]
\centering
\subfloat{\includegraphics[width=\linewidth]{Figures/new/legend.pdf}}
\vspace{-2mm}
\centering
\subfloat[{{\small \texttt{EEDI}}}]{\label{eedi-causal-nodp}\includegraphics[width=0.49\linewidth]{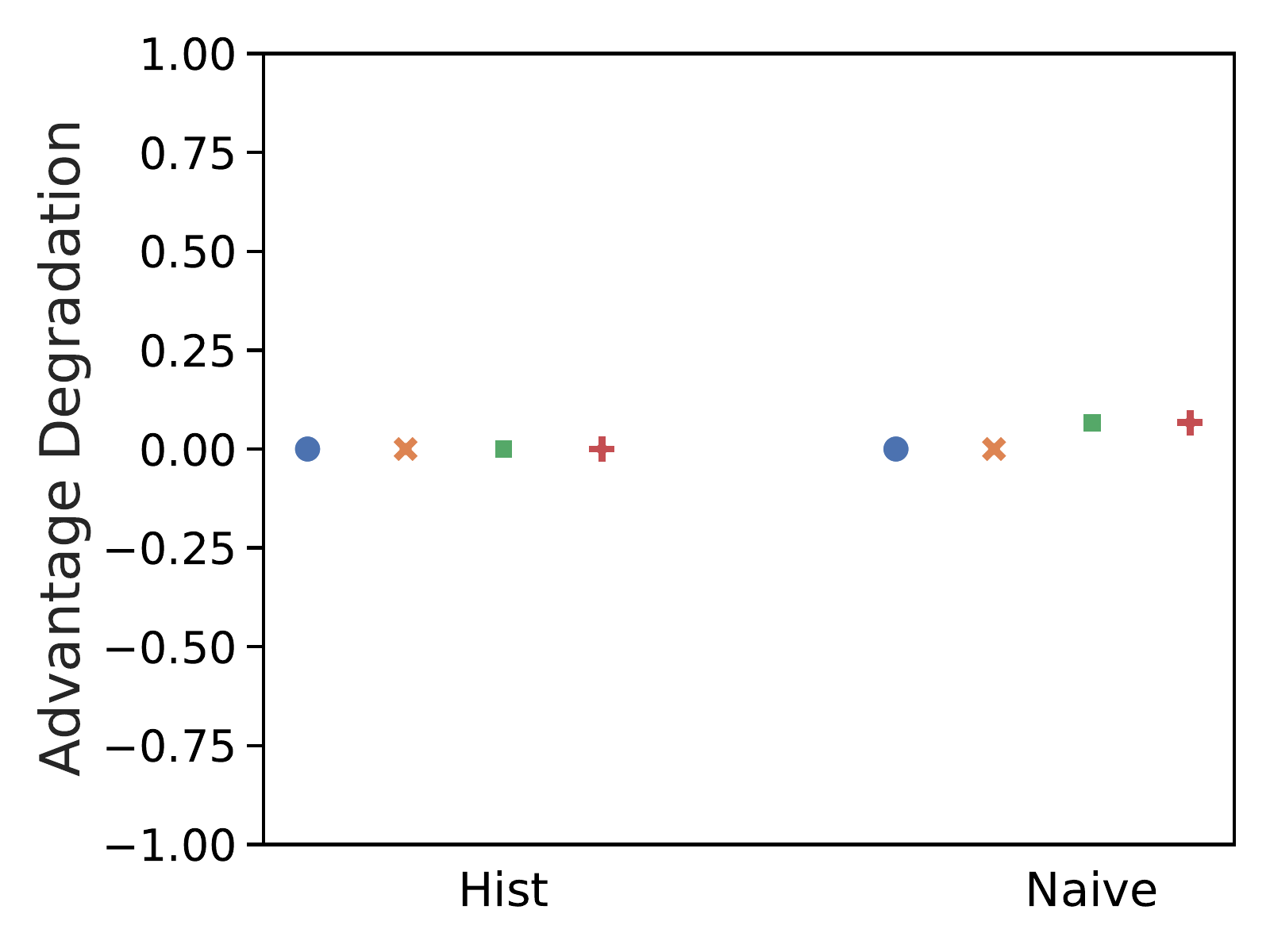}}   
\subfloat[{{\small \texttt{Pain5000}}}]{\label{pain5000-causal-nodp}\includegraphics[width=0.49\linewidth]{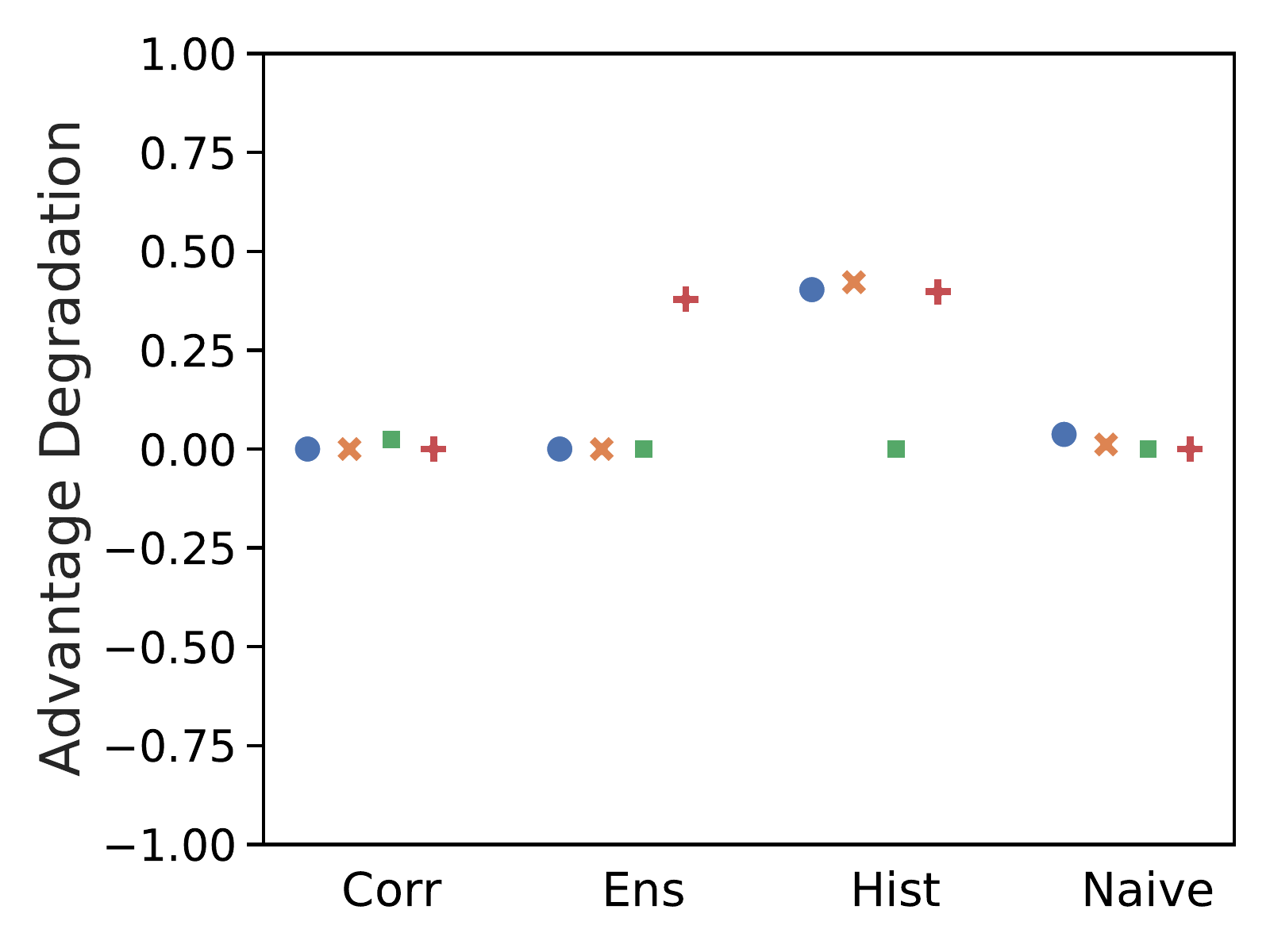}}   
\caption{{\bf Partial Causal Information \& No DP}: We plot the advantage degradation when the adversary uses a non-causal model and switches to its causal counterpart. Observe that even in the absence of DP, causal consistency by itself provides resilience against our MI adversary.  }
\label{fig:no-dp-causal}
\end{figure}

\paragraph{2. Effect of only Causality.} 

Figure~\ref{fig:no-dp-causal} shows the MI adversary's advantage when the model incorporates causal information {\em in the absence of any DP}. Observe that across both datasets, the adversary's advantage degrades (i.e., has values above zero). While this effect is moderate in the \texttt{EEDI} dataset (Figure~\ref{eedi-causal-nodp}), this effect is more pronounced in the \texttt{Pain5000} dataset (Figure~\ref{pain5000-causal-nodp}). This suggests that standalone causal information provides some privacy guarantees. We conjecture that the MI adversary uses the spurious correlation among the attributes to perform the attack. Causal learning algorithms by design eliminate any spurious correlation present in the dataset and hence, a causally consistent VAE is also able to reduce the MI adversary's advantage.

\begin{figure}
\centering
\subfloat{\includegraphics[width=\linewidth]{Figures/new/legend.pdf}}
\vspace{-2mm}
\centering
\subfloat[{{\small \texttt{EEDI}}}]{\label{eedi-causal-dp}\includegraphics[width=0.49\linewidth]{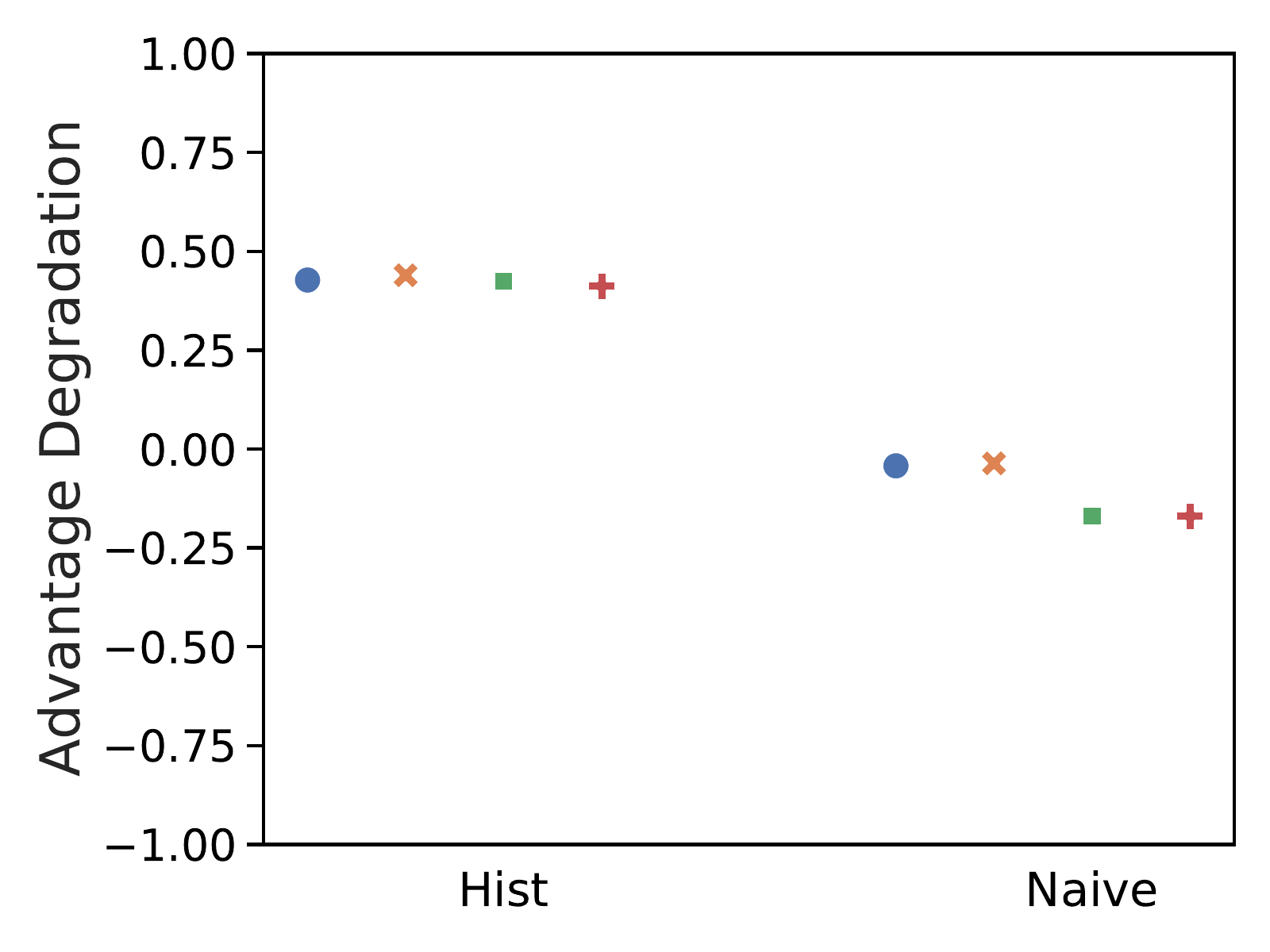}}   
\subfloat[{{\small \texttt{Pain5000}}}]{\label{pain5000-causal-dp}\includegraphics[width=0.49\linewidth]{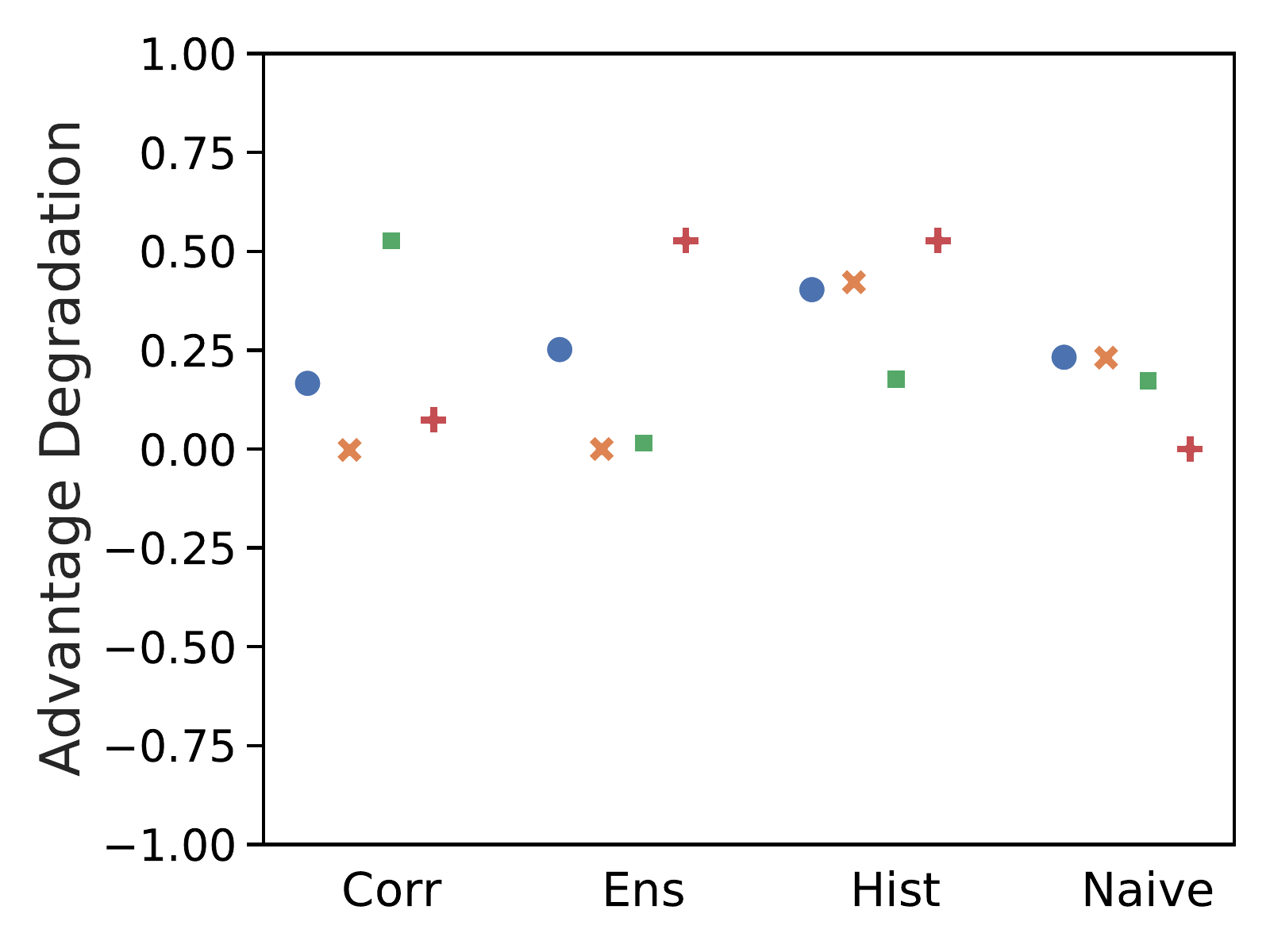}}   
\caption{{\bf Partial Causal Information \& DP}: We plot the advantage degradation when the adversary uses a non-causal model and switches to its causal counterpart. Observe that in the presence of DP \& causal consistency, the adversary's advantage is predominantly reduced. However, there are some cases where the advantage improves by using a causally consistent DP model. }
\label{fig:dp-causal}
\end{figure}

\paragraph{3. Effect of Causality with DP.} 
Figure~\ref{fig:dp-causal} shows the results for models trained using partial causal information with DP and we observe some interesting findings. 
Similar to the results with complete causal information, we observe that the combination of causality with DP is indeed effective in reducing the MI adversary's advantage for majority of the feature extractors and classifiers (shown in Figure~\ref{pain5000-causal-dp} and Figure~\ref{eedi-causal-dp} (\texttt{Hist})). However, as an exception, we see that their conjunction does increase the attacker's advantage for certain feature extractors when used with specific classifier (such as \texttt{svc} or  \texttt{svc kernel} using \texttt{Naive} features in Figure~\ref{eedi-causal-dp}).
Thus, in this setting, the behaviour is not well explained. We conjecture two reasons for this:

\begin{enumerate}
\itemsep0em
\item The membership inference adversary may exploit {\em spurious} correlations between different attributes in a record, and different records. The partial causal information provided does not entirely eliminate these correlations. 
\item The information contained in the graph is counteracted by DP training.
\end{enumerate}

\section{Future Work}
\label{sec:discussion}

Through our work, we identify several key questions that are to be analyzed in greater detail.

\begin{enumerate}
\itemsep0em
\item Our approach of incorporating causal information into the generative process has assumed that the SCG is provided. Future work is required to automate the SCG generation process from observed data and directly incorporating it in the VAE training process.
\item In our work, we show that in some scenarios, partial causal information can provide privacy amplification. However, for specific dataset and feature extractor combination, causal information increases attack efficiency used in addition to DP guarantees. A more detailed analysis is required to understand the interplay between partial causal information and DP training.
\item We also wish to understand the influence of {\em incorrect} causal information on the efficacy of MI. 
\end{enumerate}

\section{Related work}
\label{sec:related}

\paragraph{Private Data Generation:} The primary issue associated with synthetic data generation in a private manner involves dealing with data scale and dimensionality. Solutions involve using Bayesian networks to add calibrated noise to the latent representations~\citep{zhang2017privbayes,jalko2019privacy}, or smarter mechanisms to determine correlations~\citep{zhang2020privsyn}. Utilizing synthetic data generated by generative adversarial networks (GANs) for various problem domains has been extensively studied, but only few solutions provide formal guarantees of privacy~\citep{jordon2018pate,wu2019generalization,harder2020differentially,torkzadehmahani2019dp,ma2020rdp,tantipongpipat2019differentially,xin2020private,long2019scalable,liu2019ppgan}. Across the spectrum, very limited techniques are evaluated against adversaries~\citep{mukherjee2019privgan}.

\paragraph{Membership Inference:} Membership Inference (MI) is the process associated with determining whether a particular sample was used in the training of an ML model. This is a more specific form of model inversion~\citep{fredrikson2014privacy}, and was made popular by the work of~\citet{shokri2017membership}. \citet{yeom2018privacy} draw the connections to MI and overfitting, and argue that differentially private training be a candidate solution. However, most work focuses on the discriminative setting. More recently, several works propose MI attacks against generative models~\citep{stadler2020synthetic,hayes2019logan,chen2020gan,hilprecht2019monte} but offer a limited explanation as to why they are possible.

\section{Conclusions}

In this work, we propose a mechanism for private data release using VAEs trained with differential privacy. Our theoretical results highlight how causal information encoded into the training procedure can potentially {\em amplify} the privacy guarantee provided by differential privacy, without degrading utility. Our results show how knowledge of the true causal graph enables resilience against strong membership inference attacks. However, partial causal knowledge sometimes enhances the adversary. Through this body of work, we wish to better understand why membership inference is possible, and conditions (imposed by causal models) where such attacks fail.

\bibliographystyle{ICLR/iclr2021_conference}
\bibliography{ref}

\newpage
\appendix

\onecolumn
\section{Proof of Theorem 1}
\label{sec:proof}

\noindent{\bf Notation:} A mechanism $H$ takes in as input a dataset $D$ and outputs a parameterized model $f_\theta$, where $\theta$ are the parameters of the model. The model (and its parameters) belongs to a hypothesis space $\mathcal{H}$. The dataset comprises of samples, where each sample $\mathbf{x}=(x_1, \cdots, x_k)$ comprise of $k$ features. To learn the model, we utilize the empirical risk mechanism (ERM), and a loss function $\mathcal{L}$.  
The subscript of the loss function denotes what the loss is calculated over. For example $\mathcal{L}_\mathbf{x}$ denotes the loss being calculated over sample $\mathbf{x}$. Similarly, $\mathcal{L}_D$ denotes the average loss calculated over all samples in the dataset i.e. $\mathcal{L}_D = \frac{1}{|D|} \sum_{\mathbf{x} \in D} \mathcal{L}_{\mathbf{x}}$. Additionally, $\mathcal{L}_{\mathbf{x}} (f_\theta)=l(f_\theta(\mathbf{x}), f^*(\mathbf{x}))$ where $f^*$ is the oracle (responsible for generating ground truth), and $l(.,.)$ can be any loss function (such as the cross entropy loss or reconstruction loss for a generative model).

 \textbf{Data Generating Process (DGP):}  DGP $<f^*, \eta>$ is obtained as follows: $f^* = \lim_{n \to \infty} \argmin \mathcal{L}_{D}(f_\theta)$. Essentially $f^\star$
can be thought of as the {\em infinite data} limit
of the ERM learner and can be viewed 
as the {\em ground truth}. In a causal setting, the DGP for all variables/features $\mathbf{x}$ is defined as $f^*(\mathbf{x}) = (f^*_1(Pa(x_1)) + \eta_i, \cdots, f^*_n(Pa(x_n))+\eta_n)$ where $\eta_i$ are mutually, independently chosen noise values and $Pa(x_i)$ are the parents
of $x_i$ in the SCG.

\noindent{\bf Distinction between Causal and Associational Worlds:} For each feature $x_i$, we call $Pa(x_i)$ as the \textit{causal} features, and $X \setminus \{x_i, Pa(x_i)\}$ as the \textit{associational} features for predicting $x_i$. Correspondingly, the model using only $Pa(x_i)$ for each feature $x_i$ is known as the causal model, and the model using all features $X$ (including associational features) is known as the associational model. We denote the causal model learnt by ERM with loss $\mathcal{L}$ as  $f_{\theta_c}$, and the associational model learnt by ERM using the same loss $\mathcal{L}$ as $f_{\theta_a}$. Note that the hypothesis class for the models is different: $f_{\theta_c} \in \mathcal{H}_C$ and $f_{\theta_a} \in \mathcal{H}_A$, where $\mathcal{H}_C \subseteq \mathcal{H}_A$.

Like $f_{\theta_c}$, the true DGP function uses only the causal features. Assuming that the true function $f^*$ belongs in the hypothesis class $\mathcal{H}_C$, we write, $f^*=\lim_{|D|\to \infty}\arg \min_{f\in \mathcal{H}_C} \mathcal{L}_D(f)$. 

\textbf{Adversary}. Given a dataset $D$ and a model $f_\theta$,  the role of an adversary is to create a neighboring dataset $D'$ by adding a new point $\mathbf{x}'$. We assume that the adversary does so by choosing a point $\mathbf{x}'$ where the loss of $f_\theta$ is maximized. Thus, the difference of the empirical loss on $D'$ compared to $D$ will be high, which we expect to lead to high susceptibility to membership inference attacks. 


\noindent{\underline{\bf Definition: Loss-maximizing (LM) Adversary}}: Given a model $f_\theta$, dataset $D$, and a loss function $\mathcal{L}$, an LM adversary chooses a point $\mathbf{x}'$ (to be added to $D$ to obtain $D'$) as $\arg \max_{\mathbf{x}} \mathcal{L}_\mathbf{x}(f_\theta)$. Note that $\mathcal{L}_\mathbf{x}(f_\theta) = \mathcal{L}_\mathbf{x}(f_\theta(\mathbf{x}))$

\begin{tcolorbox}

\privacythm*

\end{tcolorbox}

The main steps  of our proof
are as follows:
\begin{enumerate}
\item We show that the maximum loss of a causal model is lower than or equal to maximum loss of the corresponding associational model. 
\item Using strong convexity and Lipschitz continuity of the loss function, we show how the difference in loss corresponds to the sensitivity of the learning function.
\item Finally, the  privacy budget $\varepsilon$ is a monotonic function of the sensitivity. 
\end{enumerate}

We show claim 1 separately for $n \to \infty$ (A.1) and finite $n$ (A.2) below. Then we prove claim 2 in A.3. The claim 3 follows from differential privacy literature~~\citep{dwork2014algorithmic}.

\subsection{Proof of Claim 1 ($n\to \infty$)}
As $|D| = n \to \infty$, the proof argument is that the causal model becomes the same as the true DGP $f^*$.

\noindent{\bf P1.} Given any variable $x_t$, the causal model learns a function based only on its parents, $Pa(x_t)$. 
The adversary for causal model chooses points from the DGP $<f^*,\eta>$\footnote{One should think of the DGP = $<f^*, \eta>$ as the oracle that generates labels.} s.t.,

\begin{equation}
\mathbf{x}' = \arg \max_{\mathbf{x}} \mathcal{L}_{\mathbf{x}}(f_{\theta_c}(\mathbf{x})) \text{ s.t. } \forall i \text{ \ } x_i=f_i^*(Pa(x_i)) + \eta_i 
\end{equation}

where $f_{\theta_c}=\arg \min_{f\in\mathcal{H}_C} \mathcal{L}_D(f)$. Assuming that $\mathcal{H}_C$ is expressive enough such that $f^* \in \mathcal{H_C}$, as $n=|D|\to \infty$, we can write, 
$$ \lim_{|D| \to \infty} f_{\theta_c} = \lim_{|D| \to \infty} \arg \max_{f\in\mathcal{H}_C} \mathcal{L}_D(f) = f^*$$

Therefore, the causal model is equivalent to the true DGP's function. For any target $x_i$ to be predicted, maximum error on any point is $\eta_i$ for the $\ell_1$ loss, and a function of $\eta_i$ for other losses.  Intuitively, the adversary is constrained to choose points at a maximum $\eta_i$ distance away from the causal model.

But for associational models, we have,
\begin{equation}
\mathbf{x}'' = \arg \max_{\mathbf{x}} \mathcal{L}_{\mathbf{x}}(f_{\theta_a}(\mathbf{x})) \text{ s.t. } \forall i \text{ \ } x_i=f_i^*(Pa(x_i)) + \eta_i 
\end{equation}

As $n=|D|\to \infty$, $f_{\theta_a} \neq f^*$. Thus, the adversary is less constrained and can generate points for a target $x_i$ that are generated from a different function than the associational model. For any point, the difference in the associational model's prediction and the true value is $|f_{\theta_a}(\mathbf{x})) - f^*(Pa(x_i))| +\eta_i$, which is equivalent to the loss under $\ell_1$. For a general loss function, the loss is a function of $|f_{\theta_a}(\mathbf{x}) - f^*(Pa(x_i))| +\eta_i$. Therefore, we obtain, 
\begin{equation}
    \forall i \text{ \ }\eta_i \leq |f_{\theta_a}(\mathbf{x}) - f^*(Pa(x_i))| +\eta_i \Rightarrow \max_{\mathbf{x}} \mathcal{L}_{\mathbf{x}}(f_{\theta_c}(\mathbf{x})) \leq \max_{\mathbf{x}} \mathcal{L}_{\mathbf{x}}(f_{\theta_a}(\mathbf{x}))
\end{equation}
for all losses that are increasing functions of the difference between the predicted and actual value.

\subsection{Proof of Claim 1 (Finite $n$)}

When $n$ is finite, the proof argument remains the same but we need an additional assumption on the associational model $f_{\theta_a}$ learnt from $D$. 
From learning theory~~\citep{shalev2014mltheory}, we know that the loss of $f_{\theta_c}$ will converge to that of $f^*$, while loss of $f_{\theta_a}$ will converge to loss of $f^{\infty}_{\theta_a} \neq f^*$. Thus, with high probability, $f_{\theta_c}$ will have a lower loss w.r.t. $f^*$ than $f_{\theta_a}$ and a similar argument follows as for the infinite-data case.  However, since this convergence is probabilistic and depends on the size of $n$, it is possible to obtain a $f_{\theta_c}$ that has a higher loss w.r.t. $f^*$ compared to $f_{\theta_a}$.

Therefore, rather than assuming convergence of $f_{\theta_c}$ to $f^*$, we instead rely on the property that the true DGP function $f^*$ does not depend on the associational features $x_a$.  As a result, even if the loss of the associational model is lower than the causal model on a particular point $\mathbf{x}=x_c \cup x_a$\footnote{Note that $x_a$ and $x_c$ each represent a set of features, and not a single feature.}, we can change the value of $x_a$ to obtain a higher loss for the associational model (without changing the loss of the causal model). This requires that the associational model have a non-trivial contribution from the associational (non-causal) features, sufficient to change the loss. We state the following assumption.

\noindent{\underline{\bf Assumption 1:}} If $f_{\theta_c}$ is the causal model and $f_{\theta_a}$ is the associational model, then we assume that the associational model has non-trivial contribution from the associational features. Specifically, denote $x_c$ as the causal features and $x_a$ as the associational features, such that $\mathbf{x}=x_c \cup x_a$. We define any two new points: $\mathbf{x'}=x'_c \cup x_a$ and $\mathbf{x''}=x''_c \cup x''_a$. Let us first assume a fixed value of $x_a$. The LHS (below) denotes the max difference in loss between $f_{\theta_c}$ and $f_{\theta_a}$ (i.e., change in loss between causal and associational models over the same causal features). The RHS (below) denotes  difference in loss of $f_{\theta_a}$ between $x_a$ and another value $x^*_a$, keeping $x_c$ constant (i.e., effect due to the associational features).

The inequality below can be interpreted as follows: if adversary 1 aims to find the $x'_c$ such that difference in loss between associational and causal features is highest for a given $x_a$, then there can always be another adversary 2 who can obtain a bigger difference in loss by changing the associational features (from the same $x_a$ to $x''_a$).

\begin{equation}
\begin{split}
    \exists x_a \text{\ \ \ } 
    \max_{x'_c} & \{\mathcal{L}_{\mathbf{x'}}(f_{\theta_c} (x'_c \cup x_a)) - \mathcal{L}_{\mathbf{x'}}(f_{\theta_a}(x'_c \cup  x_a))\} \\
    &\leq \min_{x''_c}  \max_{x''_a} \mathcal{L}_{\mathbf{x}''}(f_{\theta_a}(x''_c \cup x''_a)) - \mathcal{L}_{x''_c \cup  x_a}(f_{\theta_a}(x''_c \cup  x_a))
\end{split}
\end{equation}

\textit{Intuition}: Imagine that $f_{\theta_c}$ is trained initially, and then associational features are introduced to train $f_{\theta_a}$. $f_{\theta_a}$  can obtain a lower loss than $f_{\theta_c}$ by using the associational features $x_a$. In doing so, it might even change the model parameters related to $x_c$. Assumption 1 says that change in $x_c$'s parameters is small compared to the importance of the $x_a$'s parameters in $f_{\theta_a}$. For example, consider a $f^*$, $f_{\theta_c}$, $f_{\theta_a}$ to predict the value of $x_t$ such that $x_c=\{x_1\}$ and $x_a=\{x_2\}$, and consider $\ell_1$ loss.  
$$f^*=  x_1  \text{;\ \ \ } f_{\theta_c}= 2x_1 \text{;\ \ \ } f_{\theta_a} = 1.9 x_1 +\phi(x_2) $$
where $x_t=f^*(\mathbf{x}) + \eta$ and $\eta \in [-0.5, 0.5]$.  Note that without $\phi(x_2)$, the loss of the associational model is lower than the loss of causal model on any point.  However, if $x_a=x_2 \in \mathbb{R}$, then we can always set $|x_2|$ to an extreme value such that $\phi(x_2)$ overturns the reduction in loss for the associational model, without invoking Assumption 1. When $x_a$ is bounded (e.g., $x_2 \in \{0,1\}$), then Assumption 1 states that the  change in loss possible due to changing $\phi(x_2)$ is higher than the loss difference (which is $0.1$ for $\ell_1$ loss).  
If $\mathcal{H}$ was the class of linear functions and we assume $\ell_1$ loss with all features in the same range(e.g., $[0,1]$), then Assumption 1 implies that the coefficient of the associational features in $f_{\theta_a}$ is higher than the change in coefficient for the causal features from $f_{\theta_c}$ to $f_{\theta_a}$.



\begin{tcolorbox}
\noindent{\bf Lemma 1}:  Assume an LM adversary and a strongly convex loss function $\mathcal{L}$. Given a causal $ f_{\theta_c}$ and an associational model $f_{\theta_a}$ trained on dataset $D$ using ERM. The LM adversary selects two points: $\mathbf{x}'$ and $\mathbf{x}''$. Then the  {\em worst-case loss} obtained on the causal ERM model $\mathcal{L}_{\mathbf{x}'}(f_{\theta_c})$ is lower than the worst-case loss obtained on the associational ERM model $\mathcal{L}_{\mathbf{x}''}(f_{\theta_a})$ i.e.,

$$\mathcal{L}_{\mathbf{x}'}(f_{\theta_c}) \leq \mathcal{L}_{\mathbf{x}''}(f_{\theta_a})$$

which can be re-written as 

\begin{equation}
    \max_{\mathbf{x}} \mathcal{L}_{\mathbf{x}}(f_{\theta_c}) \leq \max_{\mathbf{x}} \mathcal{L}_{\mathbf{x}}( f_{\theta_a})
\end{equation}
\end{tcolorbox}

\noindent {\bf Proof}: Before we discuss the proof, let us establish a few preliminaries.




\noindent{\bf P2.} Let us write $f_{\theta_a}(\mathbf{x})=f_{\theta_a}(x_c \cup x_{a})$ as a combination of terms due to $x_c$ and $x_{a}$, where $x_c$ and $x_{a}$  are the causal features (parents) and non-causal features respectively i.e. $x_c \cup x_{a} = \mathbf{x}$, and $x_c \cap x_a = \phi$.  Let $\mathbf{x}'=x'_c \cup x'_a$ be the point chosen by the causal adversary. 

We will show that the associational adversary can always choose a point $\mathbf{x}''=x'_c \cup x''_a$ such that loss of the adversary is higher. We write, for any value $x_a$\footnote{We omit the subscript for $\mathcal{L}$ for brevity. It can be implied from context.},



\begin{equation}
\begin{split}
    \mathcal{L}(f_{\theta_a} (x'_c \cup x''_a)) 
    &=\mathcal{L}(f_{\theta_a} (x'_c \cup x''_a)) - \mathcal{L}(f_{\theta_a} (x'_c \cup x_a))  + \mathcal{L}(f_{\theta_a} (x'_c \cup  x_a))\\
    &=(\mathcal{L}(f_{\theta_a} (x'_c \cup x''_a)) - \mathcal{L}(f_{\theta_a} (x'_c \cup  x_a)))  + (\mathcal{L}(f_{\theta_a} (x'_c \cup x_a)) 
    \\ & -    \mathcal{L}(f_{\theta_c} (x'_c \cup x_a))) +
    \mathcal{L}(f_{\theta_c} (x'_c \cup x_a)) 
    \end{split}
\end{equation}

Rearranging terms, and since $\mathcal{L}(f_{\theta_c} (x'_c \cup  x'_a))=\mathcal{L}(f_{\theta_c} (x'_c \cup x_a))$ for any value of $x_a$ (causal model does not depend on associational features),
\begin{equation}
\begin{split}
    \mathcal{L}(f_{\theta_a} (x'_c \cup x''_a)) - \mathcal{L}(f_{\theta_c} (x'_c \cup x'_a)) 
    &=\underbrace{(\mathcal{L}(f_{\theta_a} (x'_c \cup x''_a)) - \mathcal{L}(f_{\theta_a} (x'_c \cup x_a)))}_\text{Term 1}  
    \\ & - \underbrace{(\mathcal{L}(f_{\theta_c} (x'_c \cup x_a)) - \mathcal{L}(f_{\theta_a} (x'_c \cup x_a))}_\text{Term 2}
    ) 
    \end{split}
\end{equation}


Now the first term is $\geq 0$ since the adversary can select $x_a''$ such that loss increases (or stays constant) for $f_{\theta_a}$. Since the true function $f^*$ does not depend on $x_a$, changing $x_a$ does not change the true function's value but will change the value of the associational model (and adversary can choose it such that loss on the new point is higher). The second term can either be positive or negative. If it is negative, then we are done. Then the LHS $> 0$. 

If the second term is positive, then we need to show that the first term is higher in magnitude than the second term. From assumption 1, let it be satisfied for some $x^\circ_a$. We know that $\mathcal{L}(f_{\theta_c} (x'_c \cup  x'_a)) = \mathcal{L}(f_{\theta_c} (x'_c \cup x^\circ_a))$ since the causal model ignores the associational features. 
\begin{equation}
\begin{split}
    \mathcal{L}(f_{\theta_c} (x'_c \cup x^\circ_a)) - \mathcal{L}(f_{\theta_a}(x'_c \cup  x^\circ_a)) \leq \max_{x_c} (\mathcal{L}(f_{\theta_c} (x_c \cup x^\circ_a)) - \mathcal{L}(f_{\theta_a}(x_c \cup  x^\circ_a))) \\ 
    \leq \min_{x_c} \max_{x^*_a} (\mathcal{L}(f_{\theta_a}(x_c \cup  x^*_a) - \mathcal{L}(f_{\theta_a}(x_c \cup  x^\circ_a)) \leq \max_{x^*_a} (\mathcal{L}(f_{\theta_a}(x'_c \cup  x^*_a) - \mathcal{L}(f_{\theta_a}(x'_c \cup  x^\circ_a))
    \end{split}
\end{equation}
Now suppose adversary chooses a point such that $x''_a=x_a^{max}$ where  $x_a^{max}$ is the arg max of the RHS above. Then Equation 7 can be rewritten as, 

\begin{equation}
\begin{split}
    \mathcal{L}(f_{\theta_a} (x'_c \cup x^{max}_a)) - \mathcal{L}(f_{\theta_c} (x'_c \cup x'_a)) 
    &=(\mathcal{L}(f_{\theta_a} (x'_c \cup x^{max}_a)) - \mathcal{L}(f_{\theta_a} (x'_c \cup x^\circ_a))) \\ & - (\mathcal{L}(f_{\theta_c} (x'_c \cup x^\circ_a)) - \mathcal{L}(f_{\theta_a} (x'_c \cup x^\circ_a))
    ) \\
    & > 0 
    \end{split}
\end{equation}
where the last inequality is due to equation 8. 

Thus, adversary can always select a different value of $\mathbf{x}=x'_c \cup x_a^{max}$
such that loss is higher than the max loss in a causal model.
$$    \mathcal{L}(f_{\theta_c} (x'_c \cup x'_a)) = 
    \max_{\mathbf{x}} \mathcal{L}_{\mathbf{x}}(f_{\theta_c}) \leq \mathcal{L}(f_{\theta_a} (x'_c \cup x^{max}_a)) \leq \max_{\mathbf{x}} \mathcal{L}_{\mathbf{x}}(f_{\theta_a})$$

\subsection{Proof of Claim 2}
\begin{tcolorbox}
\noindent{\bf Main Theorem:} Assume the existence of a dataset $D$ of $n$ samples. Further, assume a neighboring dataset is defined by adding a data point to $D$. Let $f_{\theta_c}$ and $f_{\theta_a}$ be the causal and associational models learnt using $D$, and $f_{\theta_c'}$ and $f_{\theta_a'}$ be the causal and associational models learnt using neighboring datasets $D'$ and $D''$ respectively. All models are obtained by ERM on a Lipschitz continuous,
strongly convex loss function $\mathcal{L}$. Then, the sensitivity of a {\em causal} learning function $H_C$ will be lower than that of its associational counterpart $H_A$. Mathematically speaking,
\begin{equation}
    \max_{D,D'} || \theta_c - \theta_c'|| \leq \max_{D,D'} || \theta_a - \theta_a'|| 
\end{equation}
\end{tcolorbox}

\noindent{\bf Proof:} The proof uses strongly convex and Lipschitz properties of the loss function. Before we discuss the proof, let us introduce some preliminaries.

\noindent{\bf P1.} Assume the existence of a dataset $D$ of size $n$. There are two generative models learnt, $f_{\theta{}_a}$ and $f_{\theta{}_c}$ using this dataset. Similarly, assume there is a neighboring dataset $D'$ which is obtained by adding one point $\mathbf{x}'$. Then the corresponding ERM models learnt using $D'$ are $f_{\theta{}_a'}$ and $f_{\theta{}_c'}$.

We now detail the steps of the proof.

\noindent{\bf S1.} Assume $\mathcal{L}$ is strongly convex. Then by the optimality of ERM predictor on $D$ and the definition of strong convexity, 

\begin{equation}
    \begin{split}
        &\mathcal{L}_D(f_\theta) \leq \mathcal{L}_D(\alpha f_\theta + (1-\alpha)f_{\theta'}) \\
        &\leq \alpha \mathcal{L}_D(f_\theta) + (1- \alpha) \mathcal{L}_D(f_{\theta'}) - \frac{\lambda}{2} \alpha (1-\alpha) || \theta - \theta'||^2
    \end{split}
\end{equation}
Rearranging terms, and as $\alpha \to 1$,

\begin{equation}
\begin{split}
    (1-\alpha)\mathcal{L}_D(f_\theta) \leq  (1- \alpha) \mathcal{L}_D(f_{\theta'}) - \frac{\lambda}{2} \alpha (1-\alpha) || \theta - \theta'||^2 \\
    \Rightarrow || \theta - \theta'||^2 \leq \frac{2}{\lambda} (\mathcal{L}_D(f_{\theta'}) - \mathcal{L}_D(f_\theta) ) 
    \end{split}
\label{eq:ref}
\end{equation}


\noindent{\bf S3.} Further, we can write $(\mathcal{L}_D(f_{\theta'}) - \mathcal{L}_D(f_\theta) )$ in terms of loss on $\mathbf{x}'$. 
\begin{equation}
    \begin{split}
            \mathcal{L}_{D'}(f_\theta) &= \frac{n}{n+1}\mathcal{L}_{D}(f_\theta) + \frac{1}{n+1}\mathcal{L}_{\mathbf{x}'}(f_\theta) \text{ (since $D' = D \cup \mathbf{x'}$)} \\
        & \leq  \frac{n}{n+1}\mathcal{L}_{D}(f_{\theta'}) + \frac{1}{n+1}\mathcal{L}_{\mathbf{x}'}(f_\theta) \text{ (from Equation~\ref{eq:ref}) }\\
        & \leq \frac{n}{n+1} \frac{n+1}{n}\mathcal{L}_{D'}(f_{\theta'}) - \frac{n}{n+1}\frac{1}{n} \mathcal{L}_{\mathbf{x}'}(f_{\theta'}) + \frac{1}{n+1}\mathcal{L}_{\mathbf{x}'}(f_\theta) \text{ (since $D = D' - \{\mathbf{x'}\}$)} \\
        \Rightarrow \mathcal{L}_{D'}(f_\theta) - \mathcal{L}_{D'}(f_{\theta'}) &\leq \frac{1}{n+1}  (\mathcal{L}_{\mathbf{x}'}(f_\theta) - \mathcal{L}_{\mathbf{x}'}(f_{\theta'}) )
    \end{split}
\end{equation}

\noindent{\bf S4.} Combining the above two equations, we obtain, 
\begin{equation}
    || \theta - \theta'||^2 \leq \frac{2}{\lambda} (\mathcal{L}_D(f_{\theta'}) - \mathcal{L}_D(f_\theta) ) \leq \frac{2}{\lambda(n+1)}  ( \mathcal{L}_{\mathbf{x}'}(f_\theta) - \mathcal{L}_{\mathbf{x}'}(f_{\theta'}))
\label{eq:overall}
\end{equation}

\noindent{\bf S5.} From Claim 1 above, we know that 

\begin{equation}
    \begin{split}
 \max_{\mathbf{x}} \mathcal{L}_{\mathbf{x}}(f_{\theta_c}) &\leq \max_{\mathbf{x}} \mathcal{L}_{\mathbf{x}}(f_{\theta_a}) \\
 \Rightarrow \mathcal{L}_{\mathbf{x'}}(f_{\theta_c}) &\leq \mathcal{L}_{\mathbf{x''}}(f_{\theta_a}) 
 \end{split}
\end{equation}
 where $\mathbf{x}'=\arg \max_\mathbf{x} \mathcal{L}_{\mathbf{x}}(f_{\theta_c})$ and $\mathbf{x}''$ is chosen such that $\mathbf{x'}$ and $\mathbf{x''}$ differ only in the associational features. Thus, $\mathcal{L}_{\mathbf{x}'}(f_{\theta_c'}) = \mathcal{L}_{\mathbf{x}''}(f_{\theta_c'})$. 
Also because $\mathcal H_C \subseteq \mathcal{H}_A$, the training loss of the ERM model for any $D''$ defined using $D$ and $\mathbf{x}''$ is higher for a causal model i.e., 

\begin{equation}\mathcal{L}_{\mathbf{x}'}(f_{\theta_c'}) = \mathcal{L}_{\mathbf{x}''}(f_{\theta_c'}) 
\geq \mathcal{L}_{\mathbf{x}''}(f_{\theta_a'})
\end{equation}

Therefore, we obtain, 

\begin{equation}
       \mathcal{L}_{\mathbf{x}'}(f_{\theta_c}) -  \mathcal{L}_{\mathbf{x}'}(f_{\theta_c'})= \max_\mathbf{x} \mathcal{L}_{\mathbf{x}}(f_{\theta_c}) -  \mathcal{L}_{\mathbf{x}'}(f_{\theta_c'})  
       \leq \mathcal{L}_{\mathbf{x}''}(f_{\theta_a}) -  \mathcal{L}_{\mathbf{x}''}(f_{\theta_a'})
\end{equation}

So we have now shown that the max loss difference on a point $\mathbf{x}'$ for causal ERM models trained on neighboring datasets is lower than the corresponding loss difference over $\mathbf{x}''$ for the associational models. 

\noindent{\bf S6.} Now we use the Lipschitz property, to claim,
\begin{equation}
    \mathcal{L}_{\mathbf{x}''}(f_{\theta_a})- \mathcal{L}_{\mathbf{x}''}(f_{\theta_a'}) \leq \rho ||\theta_a-\theta_a'||
\label{eq:lipschitz}
\end{equation}

\noindent{\bf S7.} Combining Equations~\ref{eq:overall} (substituting $f_{\theta_c}$) and~\ref{eq:lipschitz}, and taking max on the RHS, we get, 

\begin{equation}
\begin{split}
    \max_{D,D'}|| \theta_c - \theta_c'||^2  \leq \frac{2}{\lambda(n+1)} \max_{D,D'} \mathcal{L}_{\mathbf{x}'}(f_{\theta_c}) -  \mathcal{L}_{\mathbf{x}'}(f_{\theta_c'}) \\ \leq \frac{2}{\lambda(n+1)} \max_{D,D'} \mathcal{L}_{\mathbf{x}'}(f_{\theta_a}) -  \mathcal{L}_{\mathbf{x}'}(f_{\theta_a'})\\ \leq \frac{2\rho}{\lambda(n+1)} \max_{D,D'} ||\theta_a-\theta_a'||
\end{split}
\end{equation}

\begin{equation}
    \Rightarrow \max_{D,D'}|| \theta_c - \theta_c'||^2 \leq \frac{2\rho}{\lambda(n+1)} \max_{D,D'} ||\theta_a-\theta_a'||
\end{equation}

For $n+1 >\frac{2\rho}{\lambda}$, 
\begin{equation}
    \max_{D,D'}|| \theta_c - \theta_c'||^2 \leq  \max_{D,D'} ||\theta_a-\theta_a'||
\end{equation}

Now if $\max_{D,D'}|| \theta_c - \theta_c'||  \geq 1$, then the result follows by taking the square root over LHS. If not, we need a sufficiently large $n$ such that $n+1 >\frac{2\rho}{\lambda \max_{D,D'}|| \theta_c - \theta_c'||} $,  then we obtain,

\begin{equation}
    \max_{D,D'}|| \theta_c - \theta_c'|| \leq  \max_{D,D'} ||\theta_a-\theta_a'||
\end{equation}

\textbf{Remark on Theorem 1.} Theorem 1 depends on two key assumptions: 1) Assumption 1 that constrains associational model to have non-trivial contribution from associational (non-causal) features; and 2) a sufficiently large $n$ as shown above. When any of these assumptions is violated (e.g., a small-$n$ training dataset or an associational model that is negligibly dependent on the associational features), then it is possible that the causal ERM model has higher privacy budget than the associational ERM model.

\section{Training Details}
\label{app:training}

\subsection{Models}

\subsubsection{\texttt{Pain}}

\pgfdeclarelayer{background}
\pgfdeclarelayer{foreground}
\pgfsetlayers{background,main,foreground}

\begin{center}
\begin{tikzpicture}
\tikzstyle{scalarnode} = [circle, draw, fill=white!11,  
    text width=1.2em, text badly centered, inner sep=2.5pt]
\tikzstyle{arrowline} = [draw,color=black, -latex]
;
\tikzstyle{dasharrowline} = [draw,dashed, color=black, -latex]
;
\tikzstyle{surround} = [thick,draw=black,rounded corners=1mm]
    \node [scalarnode, fill=black!30] at (0,0) (X)   {$X_2$};
    \node [scalarnode, fill=black!30] at (-1.2, 1.5) (Y) {$X_1$};
    \node [scalarnode] at (1.2, 1.5) (Z) {$Z$};
    \path [arrowline]  (Y) to (X);
    \path [arrowline]  (Z) to (X);
    \path [arrowline]  (Z) to (Y);


    \path [dasharrowline]  (X) to[out=20,in=-80, distance=0.5cm ] (Z);
    \path [dasharrowline]  (Y) to[out=45,in=145, distance=0.5cm ] (Z);

    \node[surround, inner sep = .5cm] (f_N) [fit = (Z)(X)(Y) ] {};
\end{tikzpicture}
\end{center}

\noindent{\bf Encoder:} 
$$q_{\phi}(z, x_1|x_2) = q_{\phi_1} (z|x_1, x_2) \cdot q_{\phi_2} (x_1|x_2)$$

\noindent{\bf Decoder:} 
$$p_{\theta} (x_2, x_1, z) =  p_{\theta_1}(x_2|x_1)\cdot p_{\theta_2}(x_2|z)\cdot p_{\theta_3}(x_1|z)$$

\subsubsection{\texttt{EEDI}}

\begin{center}
\begin{tikzpicture}
\tikzstyle{scalarnode} = [circle, draw, fill=white!11,  
    text width=1.2em, text badly centered, inner sep=2.5pt]
\tikzstyle{arrowline} = [draw,color=black, -latex]
;
\tikzstyle{dasharrowline} = [draw,dashed, color=black, -latex]
;
\tikzstyle{surround} = [thick,draw=black,rounded corners=1mm]
    \node [scalarnode, fill=black!30] at (0,0) (X)   {$X_2$};
    \node [scalarnode, fill=black!30] at (-1.2, 1.5) (Y) {$X_1$};
    \node [scalarnode] at (1.2, 1.5) (Z) {$Z$};
    \path [arrowline]  (Y) to (X);
    \path [arrowline]  (Z) to (X);


    \path [dasharrowline]  (X) to[out=20,in=-80, distance=0.5cm ] (Z);

    \node[surround, inner sep = .5cm] (f_N) [fit = (Z)(X)(Y) ] {};
\end{tikzpicture}
\end{center}

\noindent{\bf Encoder:} 
$$q_{\phi}(z, x_1|x_2) = q_{\phi_1} (z| x_2)$$

\noindent{\bf Decoder:} 
$$p_{\theta} (x_2, x_1, z) = p(z) \cdot p(x_1) \cdot  p_{\theta_1}(x_2|z) \cdot p_{\theta_2}(x_2|x_1)$$

\noindent{\bf Note:} All encoders and decoders used as part of our experiments comprised of simple feed-forward architectures. In particular, these architectures have 3 layers. All embeddings generated are of size 10. We set the learning rate to be $0.001$.

\subsection{Training Parameters}

\begin{table}[H]
\small
\begin{center}
\begin{tabular}{c c c c}
\toprule
\bf Causality  & \bf Dataset & \bf Batch Size & \bf Epochs \\
\midrule
\midrule
$\times$ & \texttt{EEDI} & 200 & 1000 \\
$\checkmark$ & \texttt{EEDI} & 200 & 1000 \\ \midrule
$\times$ & \texttt{Pain1000} & 100 & 100 \\
$\checkmark$ & \texttt{Pain1000} & 100 & 100 \\ \midrule
$\times$ & \texttt{Pain5000} & 100 & 100 \\
$\checkmark$ & \texttt{Pain5000} & 100 & 100 \\ \midrule
$\times$ &\texttt{Synthetic}& 100 & 50 \\
$\checkmark$ &\texttt{Synthetic}& 100 & 50 \\
\bottomrule
\end{tabular}
\end{center}
\caption{Training parameters for our experimental evaluation.}
\label{}
\end{table}

\section{Structured Causal Graphs}
\label{sec:graphs}

\noindent{\bf SCG-1:} SCG for the \texttt{Pain} dataset. $X_1$ denotes the causes of a medical condition, and $X_2$ denotes the various conditions.
    
\begin{center}
    
\begin{tikzpicture}[->,>=stealth',shorten >=1pt,auto,node distance=1.5cm, thick,main node/.style={circle,draw,font=\sffamily\Large\bfseries}];

\tikzstyle{dasharrowline} = [draw,dashed, color=black, -latex];
                    
 \centering

  \node[main node] (1) {$X_2$};
  \node[main node] (2) [below left of=1] {$X_1$};
  \node[main node] (4) [below right of=1] {$Z$};

  \path[every node/.style={font=\sffamily\Large}]
    (4) edge node [right] {} (1)
    (2) edge node [right] {} (1)
    (4) edge node [right] {} (2);

    
\end{tikzpicture}
\end{center}

\noindent{\bf SCG-2:} SCG used by the \texttt{EEDI}. $X_2$ denotes the answers to questions and $X_1$ is the student meta data such as the year group and school.

\begin{center}

\begin{tikzpicture}[->,>=stealth',shorten >=1pt,auto,node distance=1.5cm,
                    thick,main node/.style={circle,draw,font=\sffamily\Large\bfseries}]
                    
 \centering

  \node[main node] (1) {$X_2$};
  \node[main node] (2) [below left of=1] {$X_1$};
  \node[main node] (4) [below right of=1] {$Z$};

  \path[every node/.style={font=\sffamily\Large}]
    (4) edge node [right] {} (1)
    (2) edge node [right] {} (1);
\end{tikzpicture}
\end{center}

\section{Utility Evaluation}
\label{app:utility}

\subsection{Accuracy on Original Data}

The results are detailed in Table~\ref{tab:utility-baseline}.

\begin{table}[H]
\tiny
\centering
\begin{tabular}{@{}c|ccccc|ccccc@{}}
\toprule
                 \bf Dataset  & \multicolumn{5}{c|}{\bf Non Causal} & \multicolumn{5}{c}{\bf Causal} \\ \midrule \midrule
 &  \texttt{kernel}  & \texttt{svc}   & \texttt{logistic}    & \texttt{rf}  & \texttt{knn} & \texttt{kernel}   & \texttt{svc}    & \texttt{logistic}    & \texttt{rf}  & \texttt{knn} \\ \cmidrule{2-11}
{\texttt{EEDI} }  &  86.93 &  91.24 & 88.53 & 93.8 & 88.22  &  87.38  & 91.46   & 88.95   & 93.44  & 87.92  \\
{\texttt{Pain1000}}  & 92.75   & 94.42    & 91.56   & 94.19  & 88.78  &  92.69 &  94.03  &  91.81  & 94.28  & 89.5  \\ 
{\texttt{Pain5000}}  &  95.37 &  96.53  & 97.47   & 93.03  & 92.14  & 95.48   & 96.5   &  97.39  & 92.74  & 92.19  \\ 
\bottomrule
\end{tabular}
\caption{{\bf Baseline Accuracy} calculated on the original (and not synthetic) data. Results presented in Table~\ref{tab:utility} are based on these values.}
\label{tab:utility-baseline}
\end{table}

\subsection{Privacy vs. Utility}

\begin{figure}[h]
\centering
\subfloat[\texttt{EEDI}]{\label{}\includegraphics[width=0.33\linewidth]{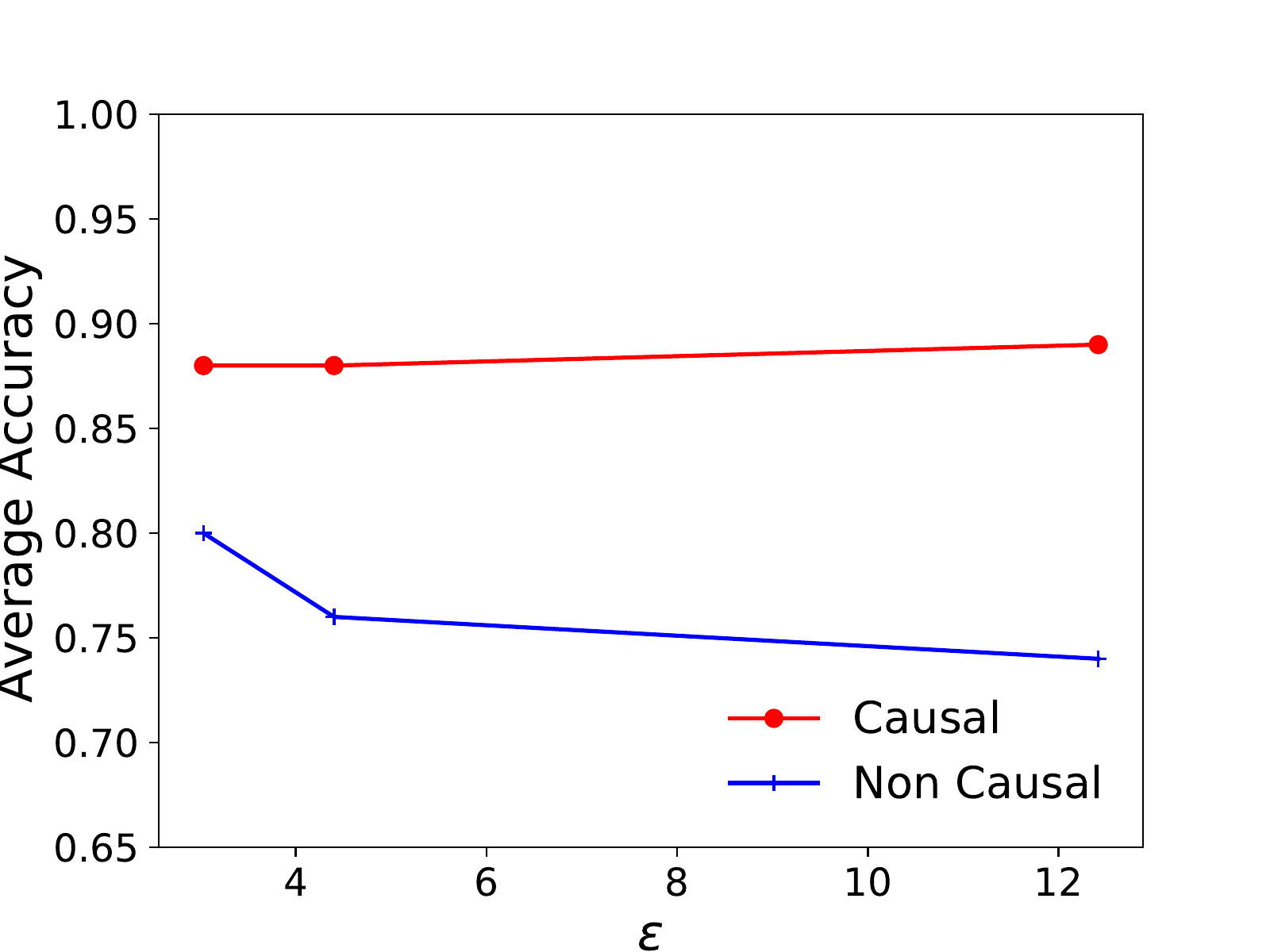}}   
\subfloat[\texttt{Pain1000}]{\label{}\includegraphics[width=0.33\linewidth]{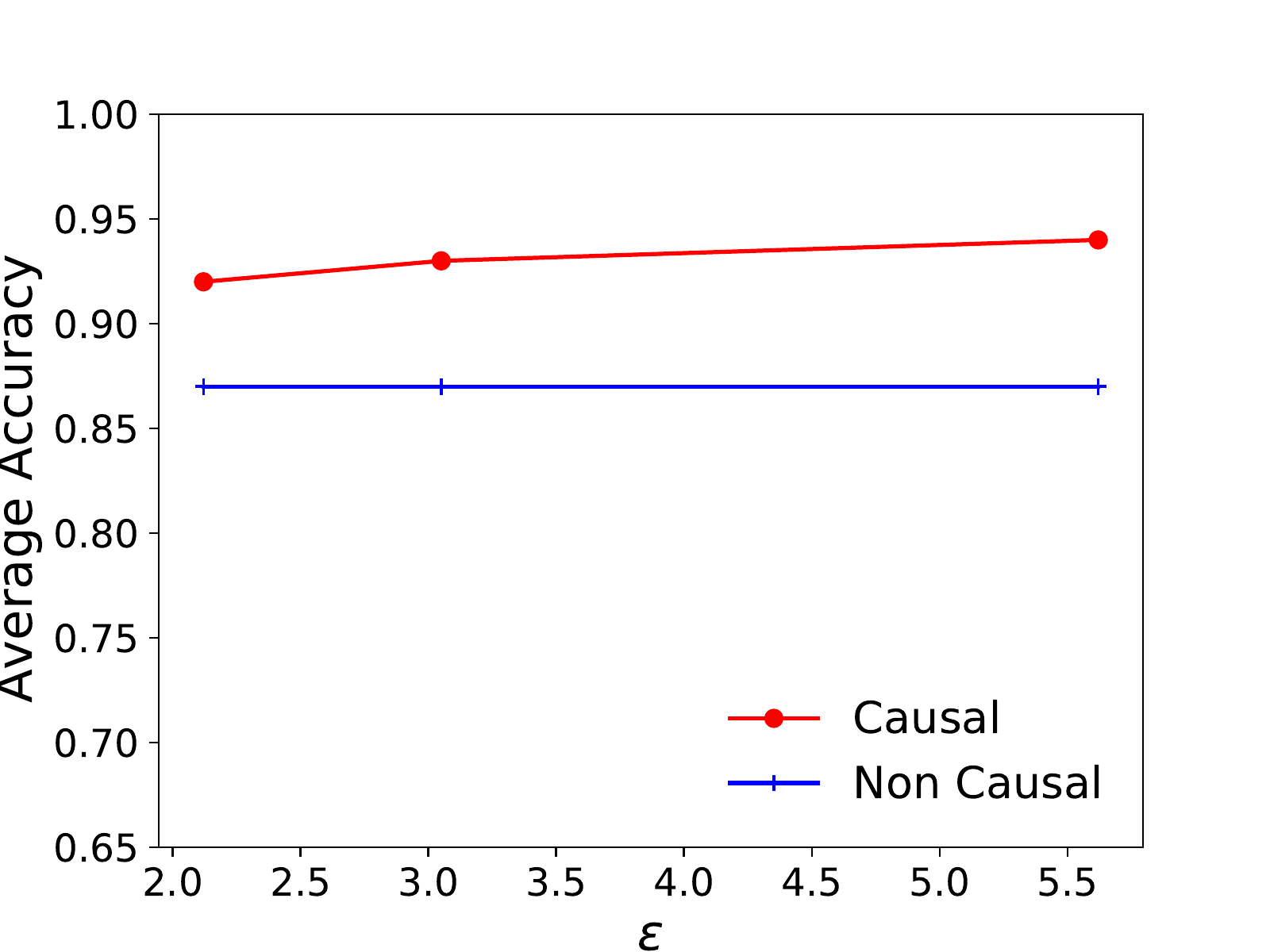}}   
\subfloat[\texttt{Pain5000}]{\label{}\includegraphics[width=0.33\linewidth]{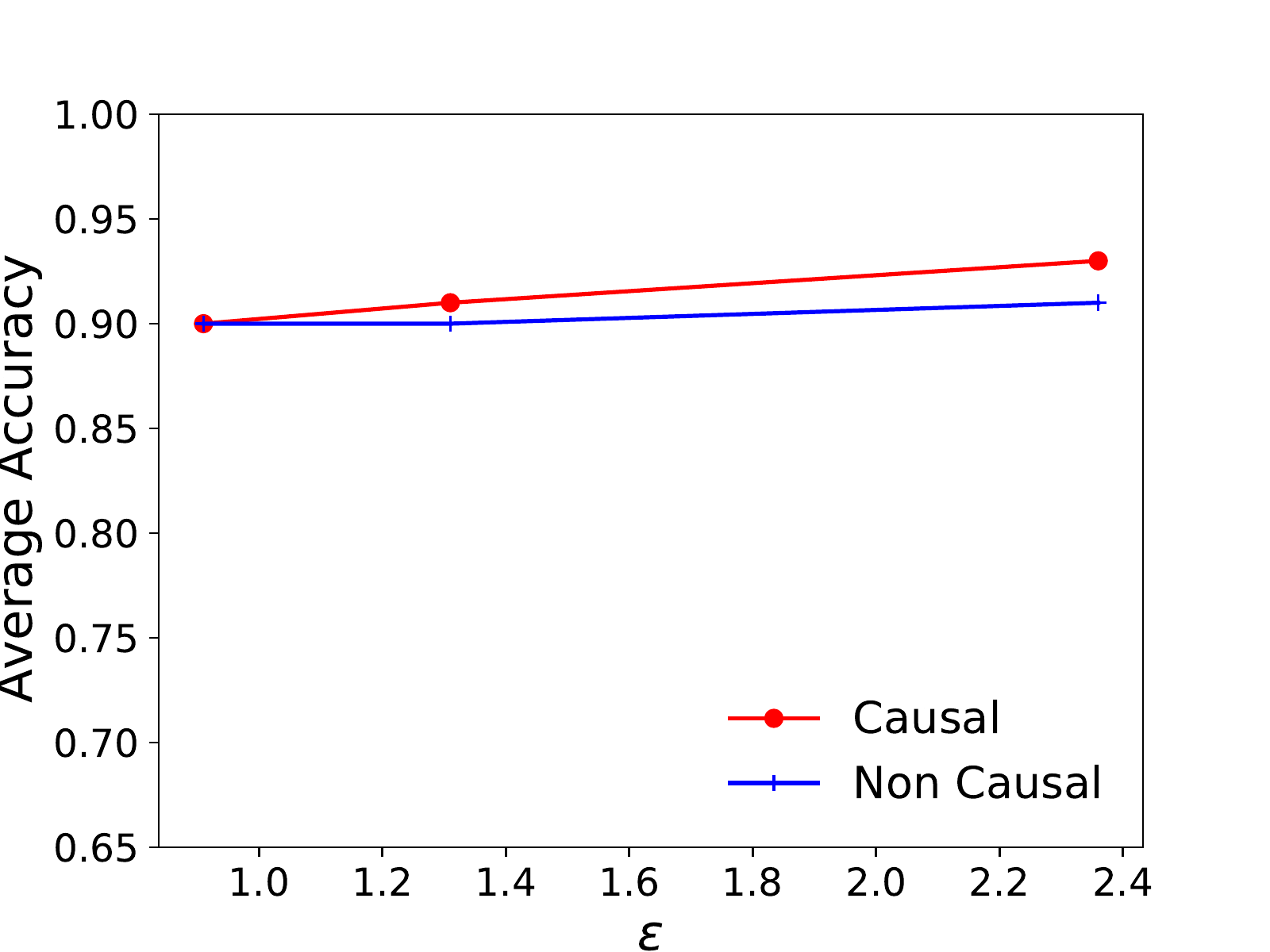}}   
\caption{{\bf Utility vs. Privacy:} Observe that causal models always outperform their associational counterparts, for the same reported value of $\varepsilon$.}
\label{fig:privutility}
\end{figure}

In Figure~\ref{fig:privutility}, we plot the utility (measured by the average accuracy across the 5 downstream prediction tasks) for both causal and associational models, for varied values of $\varepsilon$ (obtained by reducing the batch size during training). Observe that for a fixed privacy budget, the causal models always have better utility than their associational counterparts.

\subsection{Pairplots}

\begin{figure}[h]
\centering
\subfloat[Original]{\label{}\includegraphics[width=0.33\linewidth]{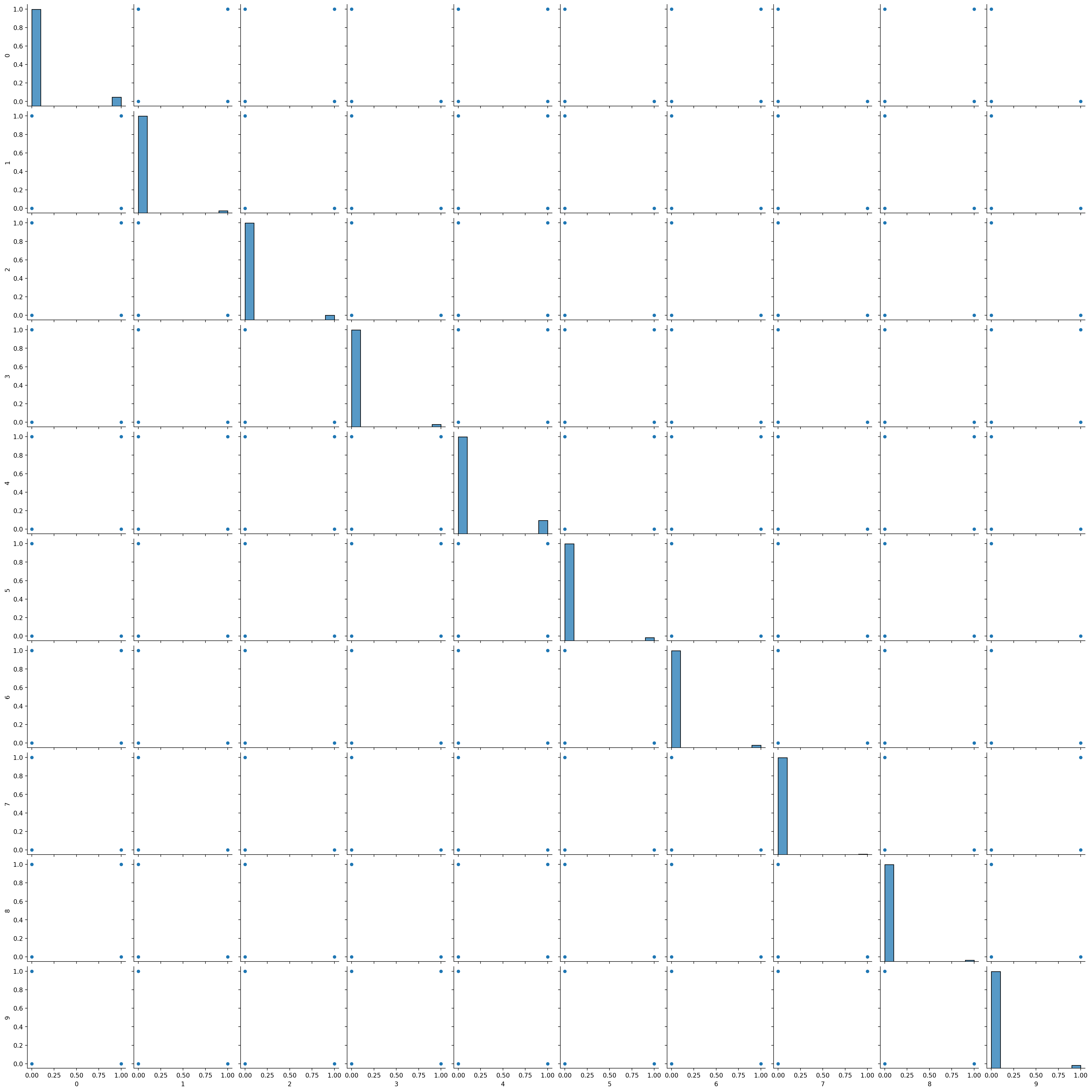}}   
\subfloat[Causal + No DP]{\label{}\includegraphics[width=0.33\linewidth]{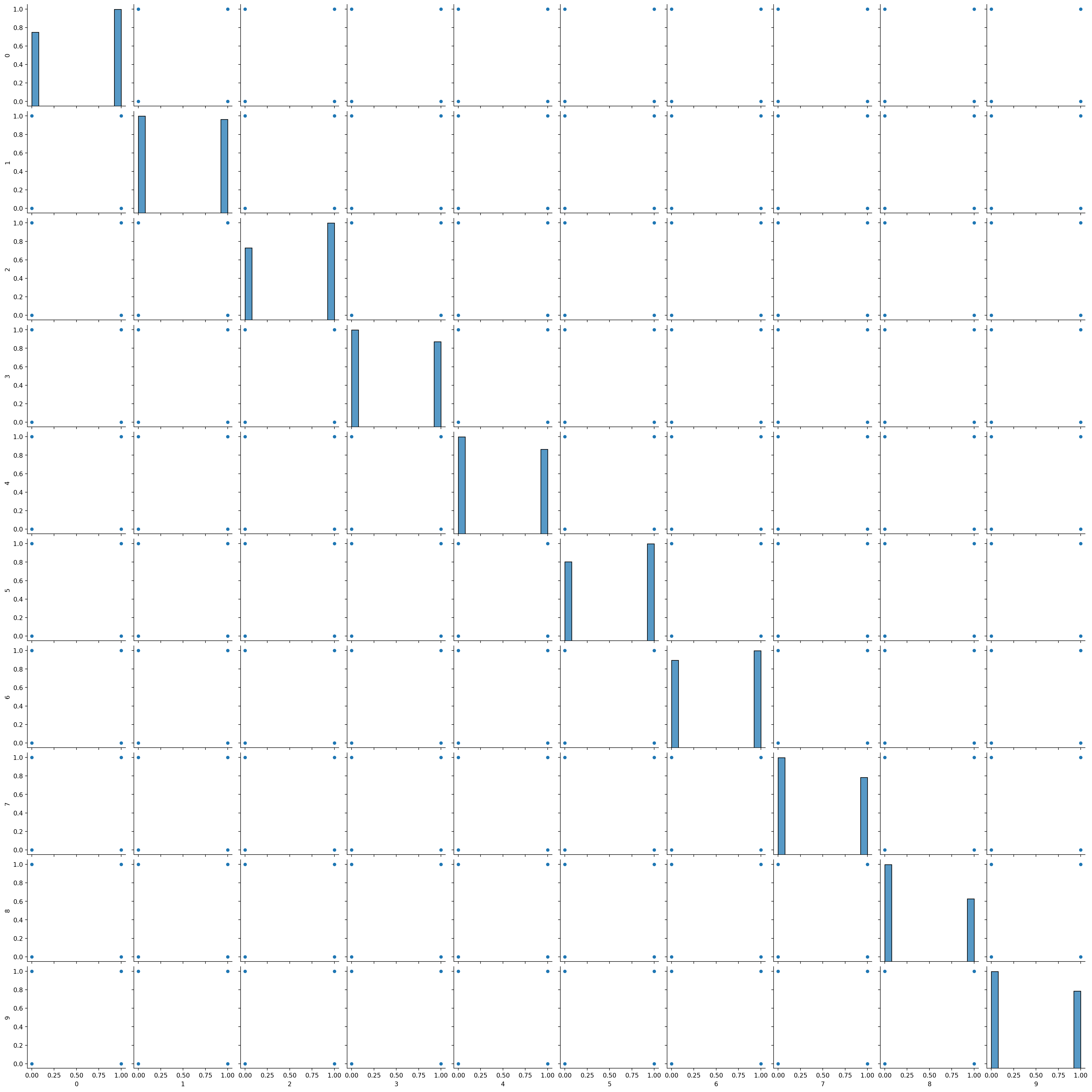}}   
\subfloat[Causal + DP]{\label{}\includegraphics[width=0.33\linewidth]{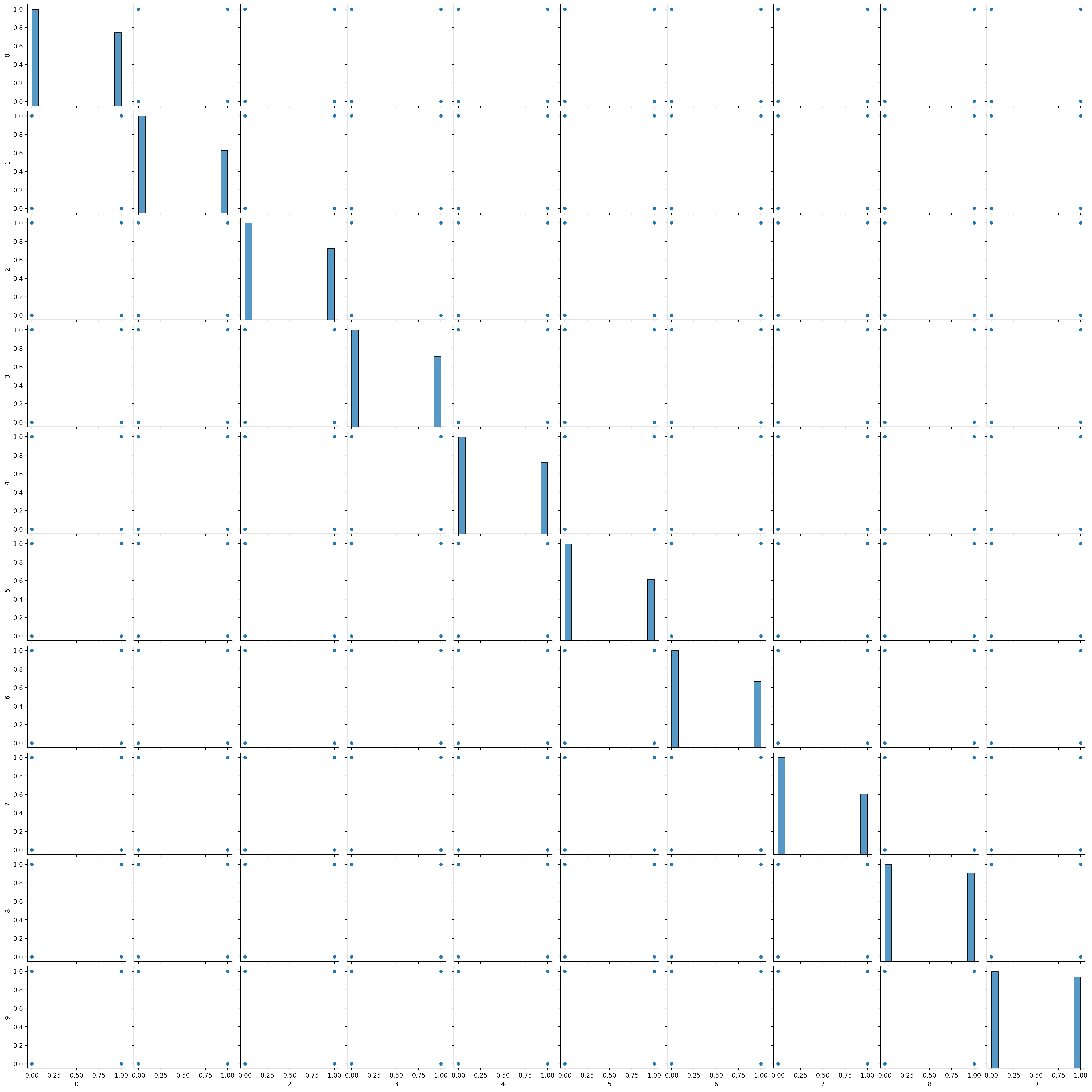}}   
\caption{\texttt{Pain5000} Dataset}
\label{fig:pairplots}
\end{figure}

\newpage
\section{Results on \texttt{Pain1000} Dataset}
\label{app:pain}

\begin{figure}[H]
\centering
\subfloat{\includegraphics[width=0.5\linewidth]{Figures/new/legend.pdf}}
\vspace{-2mm}
\centering
\subfloat[{{ Effect of only DP}}]{\label{fig:synthetic1000-non}\includegraphics[width=0.25\linewidth]{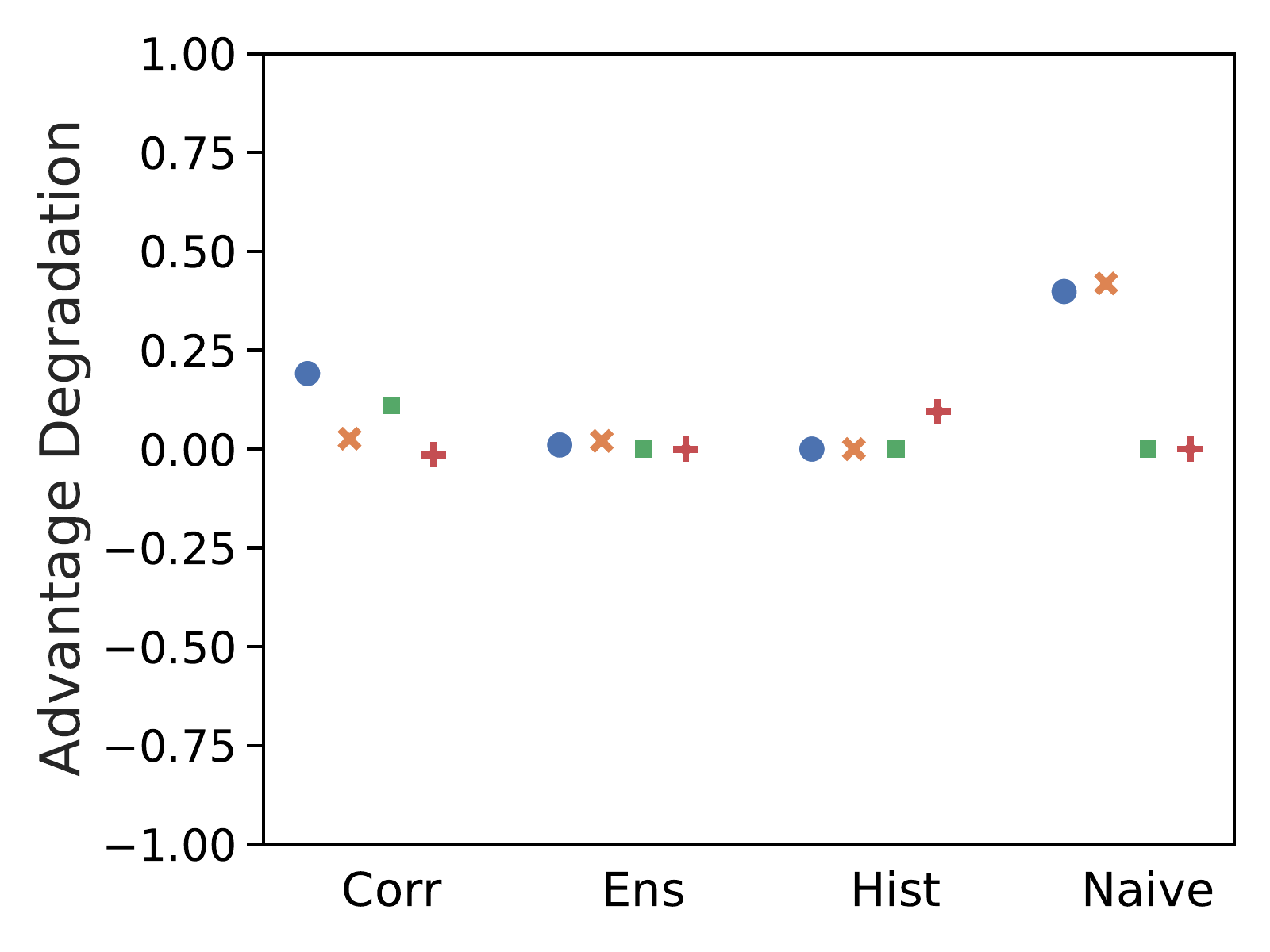}}   
\subfloat[{{Effect of only Causality}}]{\label{fig:synthetic1000-nodp}\includegraphics[width=0.25\linewidth]{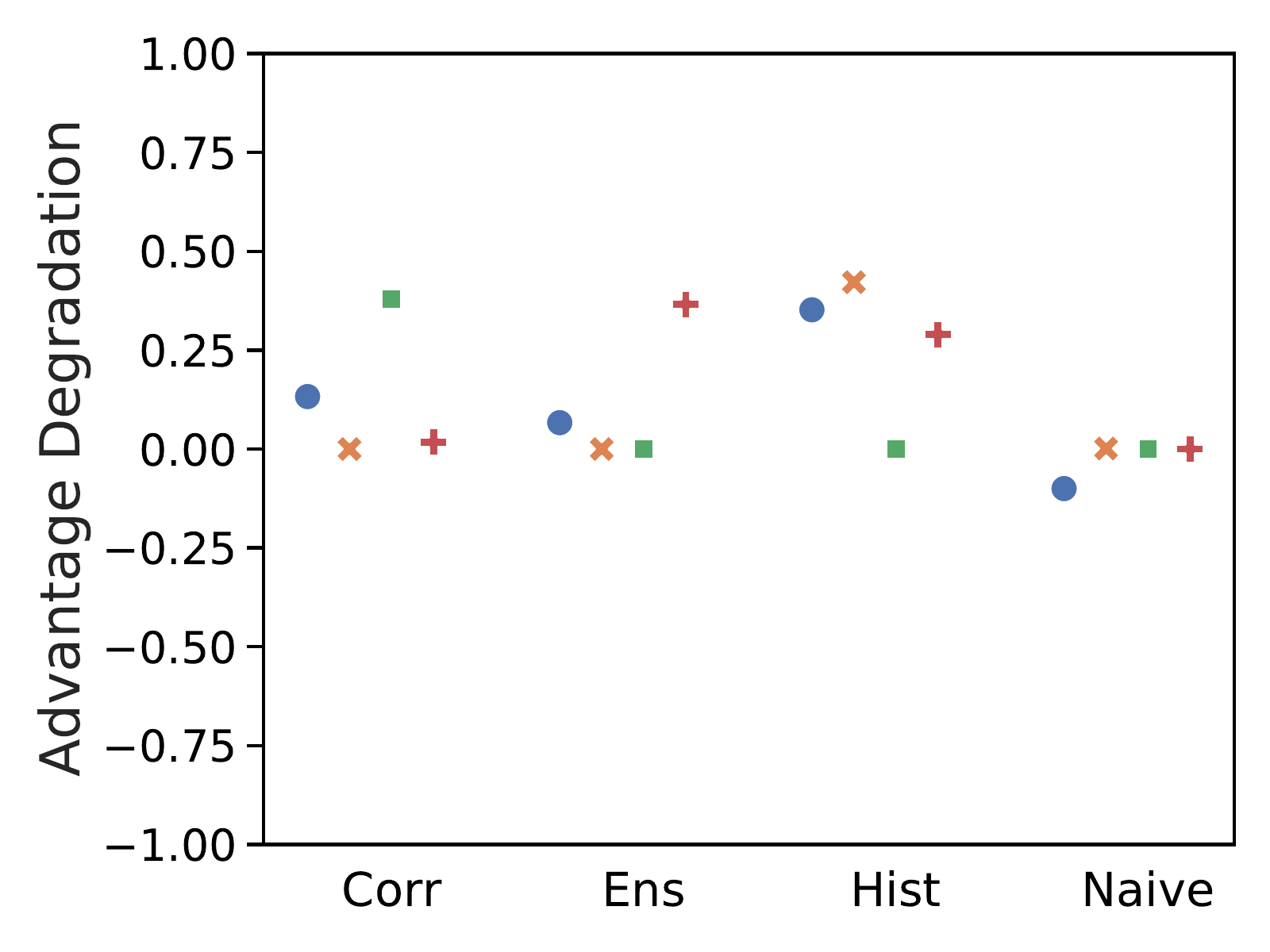}}   
\subfloat[{{Effect of causality with DP}}]{\label{fig:synthetic1000-dp}\includegraphics[width=0.25\linewidth]{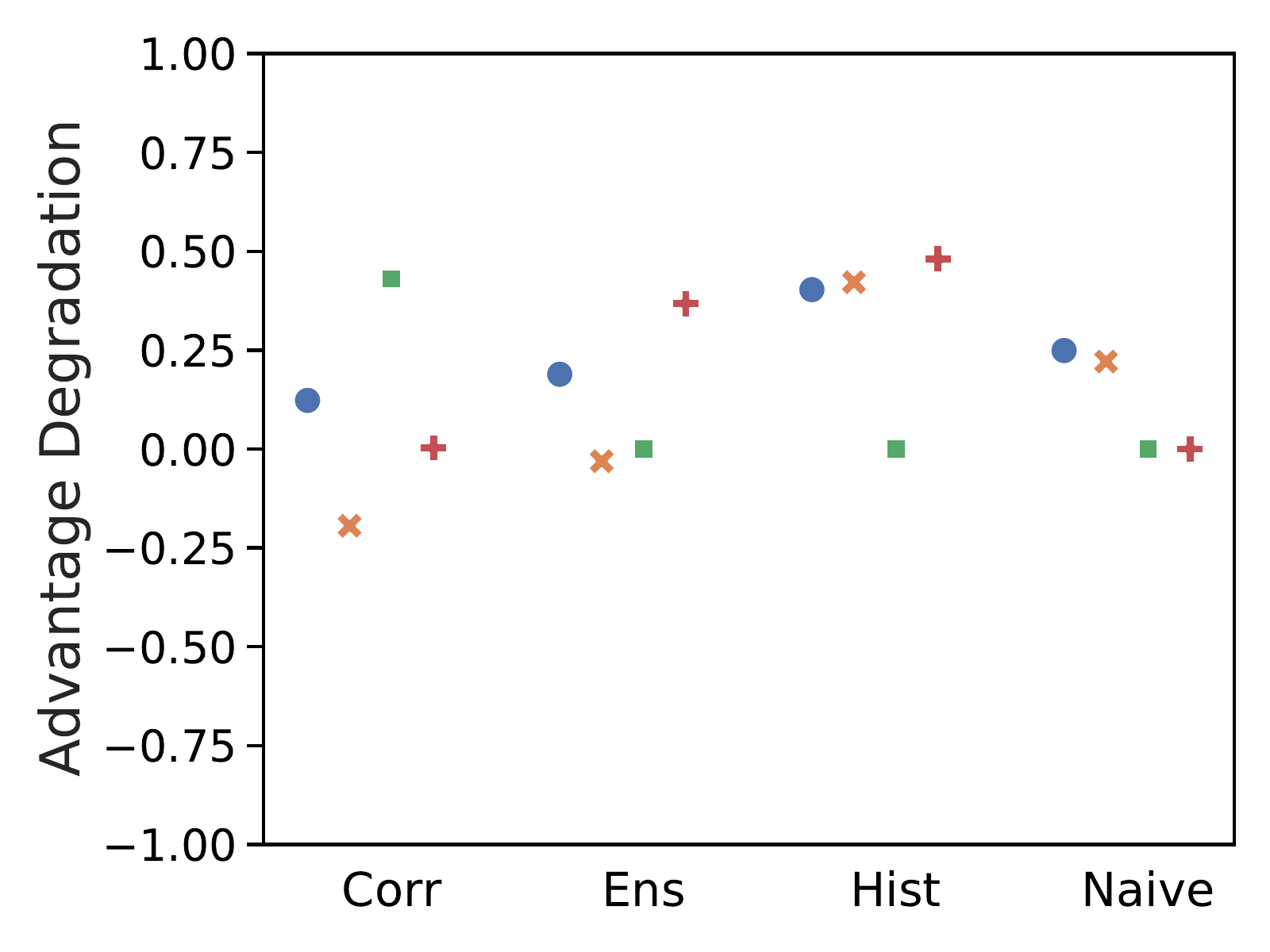}}   

\caption{{\bf Impact of DP:} Observe that in the causal scenario, DP combined with causal models enables more efficient attacks in some cases.} 
\label{fig:synthetic1000-all}
\end{figure}

In Figure~\ref{fig:synthetic1000-all}, we plot the advantage obtained by using only differential privacy (or DP), only causality and causality with DP. Values greater than zero implies that there is a reduction in adversary's advantage. We observe that DP and causality individually helps mitigate the MI adversary. But the combination of DP with causality, shown in Figure~\ref{fig:synthetic1000-dp} helps the adversary with certain classifier and feature extractor combination.

\noindent{\bf Takeaway:} As discussed in \S~\ref{sec:privacy-eval-partial}, partial causal information induces strange artifacts. Additionally, the limited amount of data contained in \texttt{Pain1000} in comparison to \texttt{Pain5000} may also contribute to some of these results.

\newpage
\section{MI Results}

{\bf PA} denotes the ability of the classifier to correctly classify train samples. {\bf NA} denotes the ability of the classifier to correctly classify test samples. The {\bf Accuracy} is a weighted combination of {\bf PA} and {\bf NA}.

\label{app:mi}

\paragraph{EEDI Non Causal:} Refer Table~\ref{tab:eedi-non-causal}.

\begin{table}[H] 
\begin{center}
\begin{tabular}{c c c c c c}
\toprule
\bf DP & \bf Extractor & \bf Attack Model & \bf Accuracy & \bf PA & \bf NA\\
\midrule
\midrule
$\checkmark$ & Naive & kernel & 79.7 &  76.97 &  82.42\\
$\checkmark$ & Naive & svc  & 79.7 &  76.97 &  82.42 \\
$\checkmark$ & Naive & random forest  & 98.18 &  96.97 &  99.39\\
$\checkmark$ & Naive & knn  & 99.7 &  99.39 &  100.0\\ \midrule
$\checkmark$ & Histogram & kernel  & 58.79 &  100.0 &  17.58\\
$\checkmark$ & Histogram & svc  & 57.58 &  60.61 &  54.55\\
$\checkmark$ & Histogram & random forest  & 56.06 &  30.3 &  81.82\\
$\checkmark$ & Histogram & knn &  57.27 &  36.36 &  78.18\\ \midrule \midrule
$\times$ & Naive & kernel  & 82.42 &  86.67 &  78.18 \\
$\times$ & Naive & svc  & 82.42 &  86.67 &  78.18\\
$\times$ & Naive & random forest  & 100.0 &  100.0 &  100.0\\
$\times$ & Naive & knn  &  100.0 &  100.0 &  100.0\\ \midrule
$\times$ & Histogram & kernel  & 100.0 &  100.0 &  100.0\\
$\times$ & Histogram & svc &  100.0 &  100.0 &  100.0\\
$\times$ & Histogram & random forest  & 100.0 &  100.0 &  100.0\\
$\times$ & Histogram & knn  & 100.0 &  100.0 &  100.0\\ 
\bottomrule
\end{tabular}
\end{center}
\caption{MI Attack accuracy results for EEDI dataset when trained without causal information.}
\label{tab:eedi-non-causal}
\end{table}

\paragraph{EEDI Causal:} Refer Table~\ref{tab:eedi-causal}

\begin{table}[H] 
\begin{center}
\begin{tabular}{c c c c c c}
\toprule
\bf DP & \bf Extractor & \bf Attack Model & \bf Accuracy & \bf PA & \bf NA\\
\midrule
\midrule
$\checkmark$ & Naive & kernel & 62.73 &  69.09 &  56.36\\
$\checkmark$ & Naive & svc & 62.73 &  69.09 &  56.36\\
$\checkmark$ & Naive & random forest & 94.55 &  90.91 &  98.18\\
$\checkmark$ & Naive & knn & 95.45 &  92.12 &  98.79\\ \midrule
$\checkmark$ & Histogram & kernel & 100.0 &  100.0 &  100.0 \\
$\checkmark$ & Histogram & svc & 100.0 &  100.0 &  100.0 \\
$\checkmark$ & Histogram & random forest & 100.0 &  100.0 &  100.0\\
$\checkmark$ & Histogram & knn & 100.0 &  100.0 &  100.0\\ \midrule \midrule
$\times$ & Naive & kernel & 89.09 &  78.18 &  100.0\\
$\times$ & Naive & svc & 89.09 &  78.18 &  100.0\\
$\times$ & Naive & random forest & 100.0 &  100.0 &  100.0\\
$\times$ & Naive & knn & 100.0 &  100.0 &  100.0\\ \midrule
$\times$ & Histogram & kernel & 100.0 &  100.0 &  100.0\\
$\times$ & Histogram & svc & 100.0 &  100.0 &  100.0\\
$\times$ & Histogram & random forest & 100.0 &  100.0 &  100.0 \\
$\times$ & Histogram & knn & 100.0 &  100.0 &  100.0\\ 
\bottomrule
\end{tabular}
\end{center}
\caption{MI Attack accuracy results for EEDI dataset when trained with causal information}
\label{tab:eedi-causal}
\end{table}

\newpage
\paragraph{Pain 1000 Causal:} Refer Table~\ref{tab:stadler-pain1000-Non}.

\begin{table}[H] 
\begin{center}
\begin{tabular}{c c c c c c}
\toprule
\bf DP & \bf Extractor & \bf Attack Model & \bf Accuracy & \bf PA & \bf NA\\
\midrule
\midrule
$\checkmark$ & Naive & kernel & 47.33 & 100.0 & 0.0\\
$\checkmark$ & Naive & svc & 47.33 & 100.0 & 0.0\\
$\checkmark$ & Naive & random forest & 57.7 & 60.44 & 55.24\\
$\checkmark$ & Naive & knn & 55.94 & 60.44 & 51.9\\ \midrule
$\checkmark$ & Histogram & kernel & 48.3 & 78.36 & 21.29\\
$\checkmark$ & Histogram & svc & 47.33 & 100.0 & 0.0\\
$\checkmark$ & Histogram & random forest & 57.82 & 100.0 & 19.91\\
$\checkmark$ & Histogram & knn & 59.7 & 81.82 & 39.82\\ \midrule
$\checkmark$ & Correlated & kernel & 99.7 & 100.0 & 99.42\\
$\checkmark$ & Correlated & svc & 51.09 & 83.99 & 21.52\\
$\checkmark$ & Correlated & random forest & 97.39 & 95.52 & 99.08\\
$\checkmark$ & Correlated & knn & 41.52 & 42.64 & 40.51\\ \midrule
$\checkmark$ & Ensemble & kernel & 61.76 & 20.49 & 98.85\\
$\checkmark$ & Ensemble & svc & 47.33 & 100.0 & 0.0\\
$\checkmark$ & Ensemble & random forest & 97.94 & 96.93 & 98.85\\
$\checkmark$ & Ensemble & knn & 81.09 & 86.3 & 76.41\\ \midrule \midrule
 $\times$ & Naive & kernel & 47.33 & 100.0 & 0.0\\
 $\times$ & Naive & svc & 47.33 & 100.0 & 0.0\\
 $\times$ & Naive & random forest & 99.58 & 99.74 & 99.42\\
 $\times$ & Naive & knn & 96.76 & 96.16 & 95.4\\ \midrule
 $\times$ & Histogram & kernel & 57.82 & 100.0 & 19.91\\
 $\times$ & Histogram & svc & 47.33 & 100.0 & 0.0\\
 $\times$ & Histogram & random forest & 57.82 & 100.0 & 19.91\\
 $\times$ & Histogram & knn & 59.7 & 81.82 & 39.82\\ \midrule
 $\times$ & Correlated & kernel & 98.24 & 99.62 & 97.01\\
 $\times$ & Correlated & svc & 62.12 & 50.58 & 72.5\\
 $\times$ & Correlated & random forest & 100.0 & 100.0 & 100.0\\
 $\times$ & Correlated & knn & 60.61 & 60.95 & 60.3\\ \midrule
 $\times$ & Ensemble & kernel & 61.7 & 72.23 & 51.78\\
 $\times$ & Ensemble & svc & 47.33 & 100.0 & 0.0\\
 $\times$ & Ensemble & random forest & 100.0 & 100.0 & 100.0\\
 $\times$ & Ensemble & knn & 82.12 & 83.99 & 80.44\\ 
\bottomrule
\end{tabular}
\end{center}
\caption{MI Attack accuracy results for Pain1000 dataset when trained with causal information}
\label{tab:stadler-pain1000-Non}
\end{table}

\newpage
\paragraph{Pain 1000 Non Causal:} Refer Table~\ref{tab:stadler-pain1000-causal}.

\begin{table}[H] 
\begin{center}
\begin{tabular}{c c c c c c}
\toprule
\bf DP & \bf Extractor & \bf Attack Model & \bf Accuracy & \bf PA & \bf NA\\
\midrule
\midrule
$\checkmark$ & Naive & kernel & 47.33 & 100.0 & 0.0\\
$\checkmark$ & Naive & svc & 47.33 & 100.0 & 0.0\\
$\checkmark$ & Naive & random forest & 79.82 & 82.33 & 77.56\\
$\checkmark$ & Naive & knn & 80.85 & 82.71 & 79.17\\ \midrule
$\checkmark$ & Histogram & kernel & 96.36 & 96.41 & 96.32\\
$\checkmark$ & Histogram & svc & 47.33 & 100.0 & 0.0\\
$\checkmark$ & Histogram & random forest & 100.0 & 100.0 & 100.0\\
$\checkmark$ & Histogram & knn & 100.0 & 100.0 & 100.0\\ \midrule
$\checkmark$ & Correlated & kernel & 100.0 & 100.0 & 100.0\\
$\checkmark$ & Correlated & svc & 94.12 & 92.83 & 95.28\\
$\checkmark$ & Correlated & random forest & 78.0 & 79.0 & 77.1\\
$\checkmark$ & Correlated & knn & 53.82 & 74.9 & 34.87\\ \midrule
$\checkmark$ & Ensemble & kernel & 98.55 & 98.46 & 98.62\\
$\checkmark$ & Ensemble & svc & 47.33 & 100.0 & 0.0\\
$\checkmark$ & Ensemble & random forest & 94.85 & 93.98 & 95.63\\
$\checkmark$ & Ensemble & knn & 100.0 & 100.0 & 100.0\\ \midrule \midrule
$\times$ & Naive & kernel & 47.33 & 100.0 & 0.0\\
$\times$ & Naive & svc & 47.33 & 100.0 & 0.0\\
$\times$ & Naive & random forest & 99.7 & 100.0 & 99.42\\
$\times$ & Naive & knn & 85.76 & 88.6 & 83.2\\ \midrule
$\times$ & Histogram & kernel & 86.79 & 90.14 & 83.77\\
$\times$ & Histogram & svc & 47.33 & 100.0 & 0.0\\
$\times$ & Histogram & random forest & 100.0 & 100.0 & 100.0\\
$\times$ & Histogram & knn & 94.91 & 94.88 & 94.94\\ \midrule
$\times$ & Correlated & kernel & 100.0 & 100.0 & 100.0\\
$\times$ & Correlated & svc & 100.0 & 100.0 & 100.0\\
$\times$ & Correlated & random forest & 100.0 & 100.0 & 100.0\\
$\times$ & Correlated & knn & 73.88 & 72.73 & 74.91\\ \midrule
$\times$ & Ensemble & kernel & 98.3 & 99.49 & 97.24\\
$\times$ & Ensemble & svc & 47.33 & 100.0 & 0.0\\
$\times$ & Ensemble & random forest & 100.0 & 100.0 & 100.0\\
$\times$ & Ensemble & knn & 88.79 & 90.27 & 87.46\\ 
\bottomrule
\end{tabular}
\end{center}
\caption{MI Attack accuracy results for Pain1000 dataset when trained without causal information}
\label{tab:stadler-pain1000-causal}
\end{table}

\newpage

\paragraph{Pain 5000 Causal:} Refer Table~\ref{tab:stadler-pain5000-Non}.

\begin{table}[H] 
\begin{center}
\begin{tabular}{c c c c c c}
\toprule
\bf DP & \bf Extractor & \bf Attack Model & \bf Accuracy & \bf PA & \bf NA\\
\midrule
\midrule
$\checkmark$ & Naive & kernel & 47.33 & 100.0 & 0.0 \\
$\checkmark$ & Naive & svc & 47.33 & 100.0 & 0.0\\
$\checkmark$ & Naive & random forest & 76.85 & 78.36 & 75.49\\
$\checkmark$ & Naive & knn & 76.79 & 78.36 & 75.37\\ \midrule
$\checkmark$ & Histogram & kernel & 47.33 & 100.0 & 0.0\\
$\checkmark$ & Histogram & svc & 47.33 & 100.0 & 0.0\\
$\checkmark$ & Histogram & random forest & 57.82 & 100.0 & 19.91 \\
$\checkmark$ & Histogram & knn & 58.7 & 81.82 & 39.82 \\ \midrule
$\checkmark$ & Correlated & kernel & 92.61 & 94.88 & 90.56\\
$\checkmark$ & Correlated & svc & 47.33 & 100.0 & 0.0\\
$\checkmark$ & Correlated & random forest & 100.0 & 100.0 & 100.0 \\
$\checkmark$ & Correlated & knn & 36.48 & 45.07 & 28.77\\ \midrule
$\checkmark$ & Ensemble & kernel & 47.33 & 100.0 & 0.0\\
$\checkmark$ & Ensemble & svc & 47.33 & 100.0 & 0.0\\
$\checkmark$ & Ensemble & random forest & 100.0 & 100.0 & 100.0 \\
$\checkmark$ & Ensemble & knn & 74.85 & 81.56 & 68.81 \\ \midrule \midrule
$\times$ & Naive & kernel & 47.33 & 100.0 & 0.0\\
$\times$ & Naive & svc & 47.33 & 100.0 & 0.0\\
$\times$ & Naive & random forest & 98.85 & 97.7 & 99.88\\
$\times$ & Naive & knn & 96.12 & 93.98 & 98.04\\ \midrule
$\times$ & Histogram & kernel & 60.24 & 38.16 & 80.09\\
$\times$ & Histogram & svc & 47.33 & 100.0 & 0.0\\
$\times$ & Histogram & random forest & 57.82 & 100.0  & 19.91 \\
$\times$ & Histogram & knn & 59.7 & 81.82 & 39.82 \\ \midrule
$\times$ & Correlated & kernel & 100.0 & 100.0 & 100.0 \\
$\times$ & Correlated & svc & 97.64 & 100.0 & 95.51 \\
$\times$ & Correlated & random forest & 100.0 & 100.0 & 100.0 \\
$\times$ & Correlated & knn & 100.0 & 100.0 & 100.0 \\ \midrule
$\times$ & Ensemble & kernel & 62.18 & 46.73 & 76.06 \\
$\times$ & Ensemble & svc & 47.33 & 100.0 & 0.0\\
$\times$ & Ensemble & random forest & 100.0 & 100.0 & 100.0\\
$\times$ & Ensemble & knn & 100.0 & 100.0 & 100.0\\ 
\bottomrule
\end{tabular}
\end{center}
\caption{MI Attack accuracy results for Pain5000 dataset when trained with causal information}
\label{tab:stadler-pain5000-Non}
\end{table}

\newpage
\paragraph{Pain 5000 Non Causal:} Refer Table~\ref{tab:stadler-pain5000-causal}.
\begin{table}[H] 
\begin{center}
\begin{tabular}{c c c c c c}
\toprule
\bf DP & \bf Extractor & \bf Attack Model & \bf Accuracy & \bf PA & \bf NA\\
\midrule
\midrule
$\checkmark$ & Naive & kernel & 47.33 & 100.0 & 0.0\\
$\checkmark$ & Naive & svc & 64.61& 60.18& 68.58\\
$\checkmark$ & Naive & random forest & 99.94& 99.87& 100.0\\
$\checkmark$ & Naive & knn & 100.0& 100.0& 100.0\\ \midrule
$\checkmark$ & Histogram & kernel & 100.0& 100.0& 100.0\\
$\checkmark$ & Histogram & svc & 65.03& 91.55& 41.2\\
$\checkmark$ & Histogram & random forest & 100.0& 100.0& 100.0\\
$\checkmark$ & Histogram & knn & 100.0& 100.0& 100.0\\ \midrule
$\checkmark$ & Correlated & kernel & 100.0& 100.0& 100.0\\
$\checkmark$ & Correlated & svc & 100.0& 100.0& 100.0\\
$\checkmark$ & Correlated & random forest & 99.76& 99.62& 99.88\\
$\checkmark$ & Correlated & knn & 53.09& 80.41& 28.54\\ \midrule
$\checkmark$ & Ensemble & kernel & 100.0& 100.0& 100.0\\
$\checkmark$ & Ensemble & svc & 48.85& 100.0& 2.88\\
$\checkmark$ & Ensemble & random forest & 100.0& 100.0& 100.0\\
$\checkmark$ & Ensemble & knn & 100.0& 100.0& 100.0\\ \midrule \midrule
$\times$ & Naive & kernel & 47.33& 100.0& 0.0\\
$\times$ & Naive & svc & 47.33& 100.0& 0.0\\
$\times$ & Naive & random forest & 100.0& 100.0& 100.0\\
$\times$ & Naive & knn & 99.82& 99.62& 100.0\\ \midrule
$\times$ & Histogram & kernel & 100.0& 100.0& 100.0\\
$\times$ & Histogram & svc & 47.33& 100.0& 0.0\\
$\times$ & Histogram & random forest & 100.0& 100.0& 100.0\\
$\times$ & Histogram & knn & 100.0& 100.0& 100.0\\ \midrule
$\times$ & Correlated & kernel & 100.0& 100.0& 100.0\\
$\times$ & Correlated & svc & 100.0& 100.0& 100.0\\
$\times$ & Correlated & random forest & 100.0& 100.0& 100.0\\
$\times$ & Correlated & knn & 100.0& 100.0& 100.0\\ \midrule
$\times$ & Ensemble & kernel & 100.0& 100.0& 100.0\\
$\times$ & Ensemble & svc & 47.33& 100.0& 0.0\\
$\times$ & Ensemble & random forest & 100.0& 100.0& 100.0\\
$\times$ & Ensemble & knn & 100.0& 100.0& 100.0\\ 
\bottomrule
\end{tabular}
\end{center}
\caption{MI Attack accuracy results for Pain5000 dataset when trained without causal information}
\label{tab:stadler-pain5000-causal}
\end{table}



\end{document}